\documentclass[10pt,twocolumn,letterpaper]{article}

\usepackage[pagenumbers]{cvpr} 

\usepackage[dvipsnames]{xcolor}

\newcommand{\methodName}{MIRAGE\xspace}
\newcommand{\methodNameExt}{Multi-view unsupervised Image generation with cRoss-Attention GuidancE\xspace}

\definecolor{cvprblue}{rgb}{0.21,0.49,0.74}
\usepackage[pagebackref,breaklinks,colorlinks,citecolor=cvprblue]{hyperref}

\def\confName{CVPR}
\def\confYear{2024}

\usepackage{amsmath}
\usepackage{amssymb}

\usepackage{graphicx}
\usepackage{amsmath}
\usepackage{amssymb}
\usepackage{booktabs}

\usepackage{multirow}
\usepackage{multicol}
\usepackage{amssymb}
\usepackage{booktabs,array}
\usepackage{xcolor}
\usepackage{amsmath}
\usepackage{bm}
\usepackage{subcaption}
\usepackage{tabularx}
\usepackage{tikz}
\usepackage{arydshln}
\usepackage{makecell}
\usepackage{graphicx}
\usepackage{tabularray}
\usepackage{xcolor}
\usepackage{adjustbox}

\title{Multi-View Unsupervised Image Generation with Cross Attention Guidance}

\author{
    Llukman Cerkezi$^*$ \
    Aram Davtyan$^*$ \
    Sepehr Sameni \
    Paolo Favaro \\
    \textnormal{Computer Vision Group, Institute of Computer Science, University of Bern, Switzerland} \\
    {\tt\small \{llukman.cerkezi, aram.davtyan, sepehr.sameni, paolo.favaro\}@unibe.ch}
}

\begin{document}

\twocolumn[{%
\renewcommand\twocolumn[1][]{#1}%
\maketitle

\setcounter{figure}{0}
\begin{center}
    \centering
    \captionsetup{type=figure}
    \includegraphics[width=\linewidth, trim=2.4cm 25cm 3cm 15cm, clip]{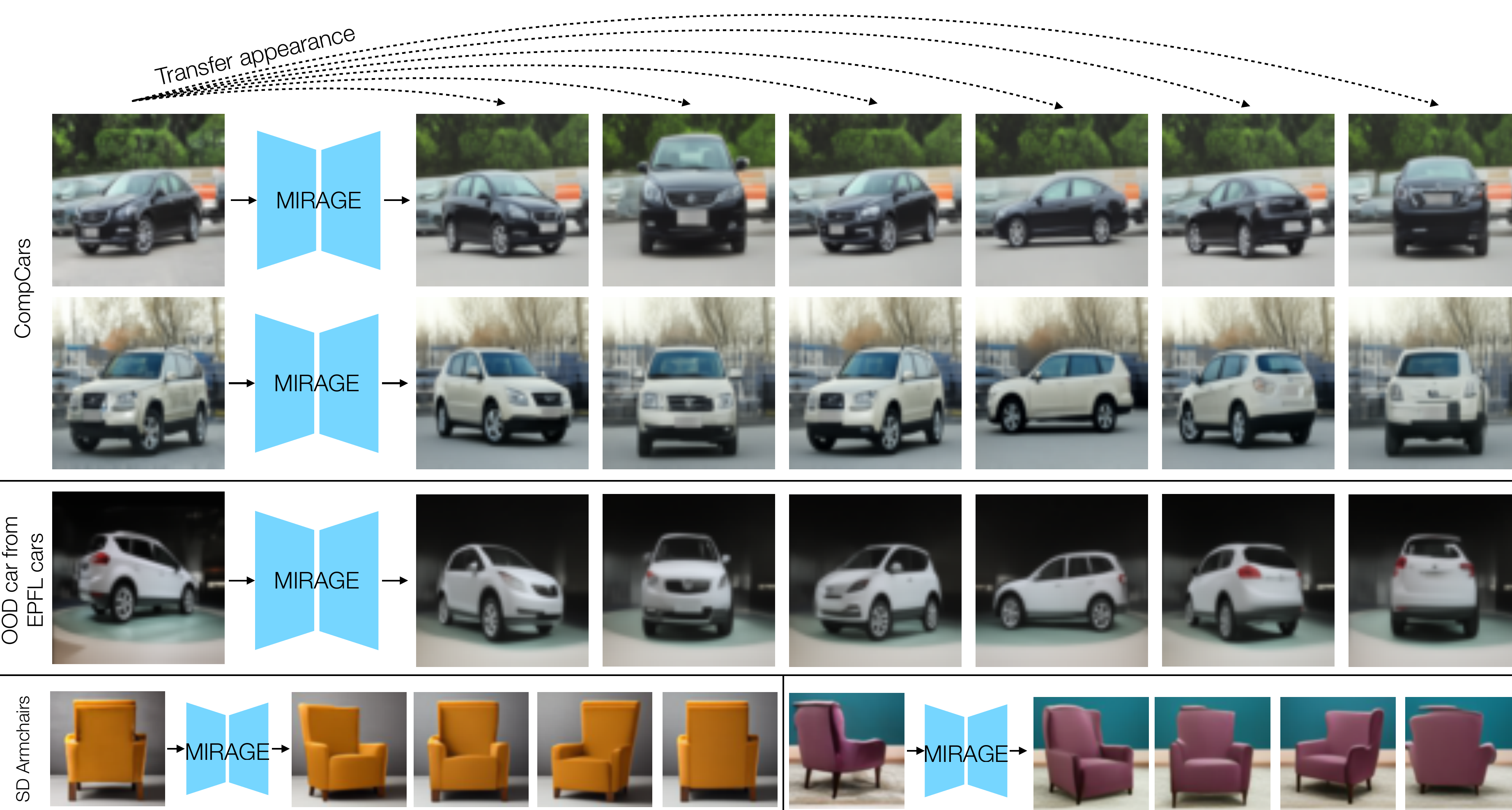}
    \captionof{figure}{
    Examples of novel views generated by \methodName on several datasets.\label{fig:teaser}}
\end{center}
}]
\def\thefootnote{*}\footnotetext{Equal contribution.}
\begin{abstract}
The growing interest in novel view synthesis, driven by Neural Radiance Field (NeRF) models, is hindered by scalability issues due to their reliance on precisely annotated multi-view images. Recent models address this by fine-tuning large text2image diffusion models on synthetic multi-view data. Despite robust zero-shot generalization, they may need post-processing and can face quality issues due to the synthetic-real domain gap. This paper introduces a novel pipeline for unsupervised training of a pose-conditioned diffusion model on single-category datasets. With the help of pretrained self-supervised Vision Transformers (DINOv2), we identify object poses by clustering the dataset through comparing visibility and locations of specific object parts. The pose-conditioned diffusion model, trained on pose labels, and equipped with cross-frame attention at inference time ensures cross-view consistency, that is further aided by our novel hard-attention guidance. Our model, \methodName, surpasses prior work in novel view synthesis on real images. Furthermore, \methodName is robust to diverse textures and geometries, as demonstrated with our experiments on synthetic images generated with pretrained Stable Diffusion.
\end{abstract}    
\section{Introduction}
\label{sec:intro}

In recent years, the field of novel view synthesis has gained significant interest, primarily due to advancements in Neural Radiance Field (NeRF) models~\cite{2020nerf,instantNGP,2021mipnerf,Nerf++}. While NeRFs have demonstrated remarkable capabilities in generating novel views, they require  hundreds of images precisely annotated with exact camera poses. This makes these models difficult to scale efficiently in practice.

To address this challenge, an alternative approach has emerged with the development of diffusion models~\cite{Ho2020DenoisingDP}. Specifically, a large conditional diffusion model is trained on a substantial multi-modal dataset of real images (e.g. Stable Diffusion~\cite{rombach2021highresolution}). This model is further fine-tuned on synthetic datasets containing multi-view images with associated camera poses to enable novel view synthesis~\cite{zero_1_to_3,Shi2023Zero123AS,yang2023consistnet}. Despite exhibiting robust zero-shot generalization, these models need additional post-processing at inference time, which involves object segmentation to remove the background. Moreover, the quality of generated novel views might still be far from satisfactory due to the domain gap between synthetic and real data. However, fine-tuning these models with real data remains unfeasible due to the impracticality and cost associated with obtaining accurate pose annotations for diverse object categories. In this work we tackle this challenge by proposing a novel pipeline based on training a pose-conditioned diffusion model on a single-category dataset in a fully unsupervised manner.

Our proposed approach starts by identifying the variety of different object poses represented in the dataset using recently developed self-supervised Vision Transformers (ViTs~\cite{vit}), namely DINOv2~\cite{dinov2}. DINOv2 has demonstrated the ability to capture semantic relationships among object parts in images of similar or related categories. Our method utilizes this capability by comparing the visibility and the location of specific object parts in images, effectively clustering the original dataset into groups that share the same pose/orientation of objects.

Following this, we train a pose-conditioned diffusion model using the pose labels obtained in the first stage. Once trained, this diffusion model allows the generation of images featuring objects in diverse and controlled poses. However, these generated images do not yet form multi-view sequences. To impose consistency across images with different poses, we employ the cross-frame attention procedure~\cite{text2video-zero}. This procedure facilitates communication of self-attention keys and values in the denoising backbone of the diffusion from the reference pose to the others. Additionally, we introduce hard-attention guidance to enhance the quality of the generated views and ensure further consistency.

We refer to our model as \methodName for \methodNameExt. To evaluate our methodology, we conduct tests on a real dataset of car images. Our results demonstrate that the generated novel views not only surpass those produced in prior work but also exhibit sufficient quality for reconstructing an explicit 3D. Furthermore, to showcase the robustness of \methodName across various categories in the absence of diverse real datasets, we train it on datasets generated with a pretrained Stable Diffusion model~\cite{rombach2021highresolution}.

Our contributions can be summarized as follows:

\begin{itemize}
    \item A novel way to discover the set of poses represented in a single-category dataset;
    \item A pose-conditioned diffusion model that is capable of generating images of objects in predefined poses;
    \item An adaptation of cross-frame attention to novel view synthesis field with our novel hard-attention guidance;
    \item A dataset of synthetic single-category images generated with Stable Diffusion to show robustness of our method to different textures and geometries.
\end{itemize}

\section{Literature Review}
\label{sec:literature}

\noindent\textbf{NeRF-based Novel view synthesis.} Neural radiance fields (NeRF)~\cite{2020nerf} have proven to be a robust framework for generating novel views. 
Despite achieving notable success \cite{2021plenoxels, 2021mipnerf, mipnerf360, TensoRF2022ECCV, instantNGP}, these models face challenges when tasked with generating novel views from a limited number of input images. 
This limitation arises from the original method's optimization, which is scene-specific.
One strategy to address this limitation involves pretraining the model across multiple scenes using synthetic data to acquire a scene prior. 
During inference, a sparse set of views, or even a single view, is then used to generate novel views \cite{2020pixelnerf, grf2020, SRF2021, 2021mvsnerf, 2021rematasICML21, lin2023visionnerf, 2023NerfDiff}. 
Note that, their inference is reliant on pretraining with multiview images and corresponding pose labels. 
Moreover, the quality diminishes when the target view significantly deviates from the input views, resulting in blurry generated views.
For this, they primarily focus on two cases 1) generating narrow views \cite{2021mvsnerf, SRF2021, 2021rematasICML21}, and 2) generating $360^\circ$ views where the input image is masked and synthetic, and the input views cover almost all the texture (from the top view) \cite{2020pixelnerf, 2023NerfDiff, lin2023visionnerf, grf2020}. 
In contrast, our work tackles a more challenging task by aiming to generate $360^\circ$ views for real images with background and without supervised training.

\noindent\textbf{Diffusion-based Novel View Synthesis.} In pioneering work, 3DiM~\cite{3dim} developed a pose conditional image-to-image diffusion model, training the network by providing the source view and relative pose to predict the target view. 
However, a significant limitation is its reliance on synthetic images, as the diffusion models are exclusively trained on synthetic datasets.
Addressing this limitation, Zero-1-to-3 \cite{zero_1_to_3} proposed fine-tuning the pre-trained multi-modal diffusion model, Stable Diffusion \cite{rombach2021highresolution}, on a large-scale synthetic dataset, Objaverse \cite{objaverse}. 
While this fine-tuning allows zero-shot generalization to out-of-distribution datasets and in-the-wild images, our attempts to generate novel views for real cars, specifically from a \textit{side} view, revealed challenges in hallucinating novel parts for other views and achieving multiview consistency.
Several works, including Consistent-1-to-3 \cite{ye2023consistent1to3}, Consistent123~\cite{weng2023consistent123}, and MVDREAM \cite{2023mvdream}, aimed to enhance multiview consistency in Zero-1-to-3. 
These works share a common methodology of modifying self-attention layers in the U-Net~\cite{Ronneberger2015UNetCN} during training, initially proposed for generating temporally consistent videos \cite{wu2023tune, text2video-zero, hong2023large} and consistent image synthesis and editing \cite{cao_2023_masactrl}. 
Specific modifications include epipolar guided attention~\cite{ye2023consistent1to3}, shared self-attention~\cite{weng2023consistent123}, and 3D attention~\cite{2023mvdream}.
In contrast, our approach differs in two key aspects. Firstly, we adjust the self-attention layer only during inference, not during training. Secondly, our novel pipeline does not require multi-view images with corresponding poses, demonstrating the ability to train with approximate and noisy discrete pose labels. 


\noindent\textbf{Single-view 3D Reconstruction.}
In contrast to novel view synthesis, single-view 3D reconstruction methods primarily focus on predicting the 3D geometry and appearance beyond the visible, rather than generating novel views. 
Common representations for 3D include mesh-based approaches \cite{cmrKanazawa18, ucmrGoel20, simoni2021multi, monnier2022unicorn}, volumetric methods \cite{Girdhar16b, choy20163d, henzler2019platonicgan, ye2021shelf}, and neural implicit 3D representations \cite{Pavllo_2023_CVPR, lin2020sdfsrn}.
A shared characteristic among these methods is their reliance on mask supervision to predict the object's shape, with the exception of Monnier \etal~\cite{monnier2022unicorn}. 
Monnier \etal achieve this by enforcing consistency between images of different object instances, ensuring similarity in shape or texture and utilizing multi-stage progressive training. 
Although they generate reasonably good 3D meshes, the texture tends to be coarse, and the training process is cumbersome due to multi-stage progressive training.
In our approach, we aim to produce more detailed and high-quality renderings of novel views while maintaining consistency. Furthermore, our method is easier and simpler to train, eliminating the need for any supervision.

\noindent\textbf{3D-aware Generative Models.}
With the development of NeRF~\cite{2020nerf}, many 3D-aware generative adversarial networks having radiance fields as a representation are proposed \cite{Schwarz2020NEURIPS,zhang2022multiview,epigraf, GIRAFFE, Chan2021_eg3d, xue2022giraffehd}.
Although they generate high-quality outputs, they often lack multiview consistency (shape and texture change when the camera moves) and are sensitive to prior pose estimation. 
This is partially due to 2D upsampler that eases the training and lack of explicit 3D supervision. 
In contrast, diffusion-based 3D-aware generative models have demonstrated superiority over GAN-based counterparts~\cite{karnewar2023holodiffusion, karnewar2023holofusion}, especially in generating $360^{\circ}$ views. However, a significant drawback is their dependence on multiview datasets with known poses, such as Co3Dv2~\cite{co3dv2}, for training.
In our work, we also show that we can obtain 3D-aware diffusion models without requiring multiview dataset.
It is noteworthy that diffusion models offer an advantage over GAN-based models in terms of inversion, making it easier to generate query input images back, thus enabling them to be also utilized for novel views. 

\section{Method}
\label{sec:method}

Let ${\cal D}$ be a collection of RGB-images capturing objects of the same category in different orientations. That is, ${\cal D} = \{ x^{(i)}\}_{i = 1}^{N}$, where $x^{(i)} \in \mathbb{R}^{3 \times H \times W}$. In this work we assume that each $x^{(i)}$ is an image of a single salient object. Furthermore, under general assumptions, ${\cal D}$ is not necessarily a multi-view dataset, i.e. it does not have to contain multiples views of the same instance. The task of novel view synthesis from a single reference image is to generate a set of images of the object in the reference frame in different orientations/poses that are consistent with the reference.

Since synthesising novel views is ambiguous due to occlusions, we propose a probabilistic approach to this task. Let ${\cal P}$ be a finite set of poses/orientations that the objects in ${\cal D}$ can have, with $|{\cal P}| = m$. Given a reference image $x_{\text{ref}}$ the task is to generate a set of images ${x_{p_1}, \dots, x_{p_m}}$, such that $x_{p_j}$ captures the object in $x_{\text{ref}}$ in the pose $p_j \in {\cal P}$. To this end, we model the following conditional distribution

\begin{align}\label{eq:cond_dist}
    p(x_{p_1}, \dots, x_{p_m} \;|\; x_{\text{ref}} ) = \prod_{j = 1}^m p(x_{p_j} \;|\; x_{\text{ref}} ).
\end{align}

Thus, each $x_{p_j}$ is sampled independently from the other views. For the separate single view conditional distributions in~\ref{eq:cond_dist} we train a pose-conditioned diffusion~\cite{Ho2020DenoisingDP} and resolve the conditioning on $x_{\text{ref}}$ through sharing keys and values in the self-attention layers~\cite{Vaswani2017AttentionIA} of the denoising U-Net~\cite{Ronneberger2015UNetCN} backbone. Recently it has been shown that this technique ensures consistency across generated images in such applications as video generation~\cite{text2video-zero} and style transfer~\cite{Alaluf2023CrossImageAF}. Indeed, although the independent sampling of the generated views may introduce inconsistencies, in the experiments section, we demonstrate empirically that the novel views obtained with our method are sufficiently good for 3D reconstruction.

\begin{figure*}
    \centering
    \begin{tabular}{@{}c:c@{}}
    \begin{subfigure}[a]{0.37\linewidth}
        \centering
        \includegraphics[width=\linewidth, trim=15cm 4cm 15cm 4cm, clip]{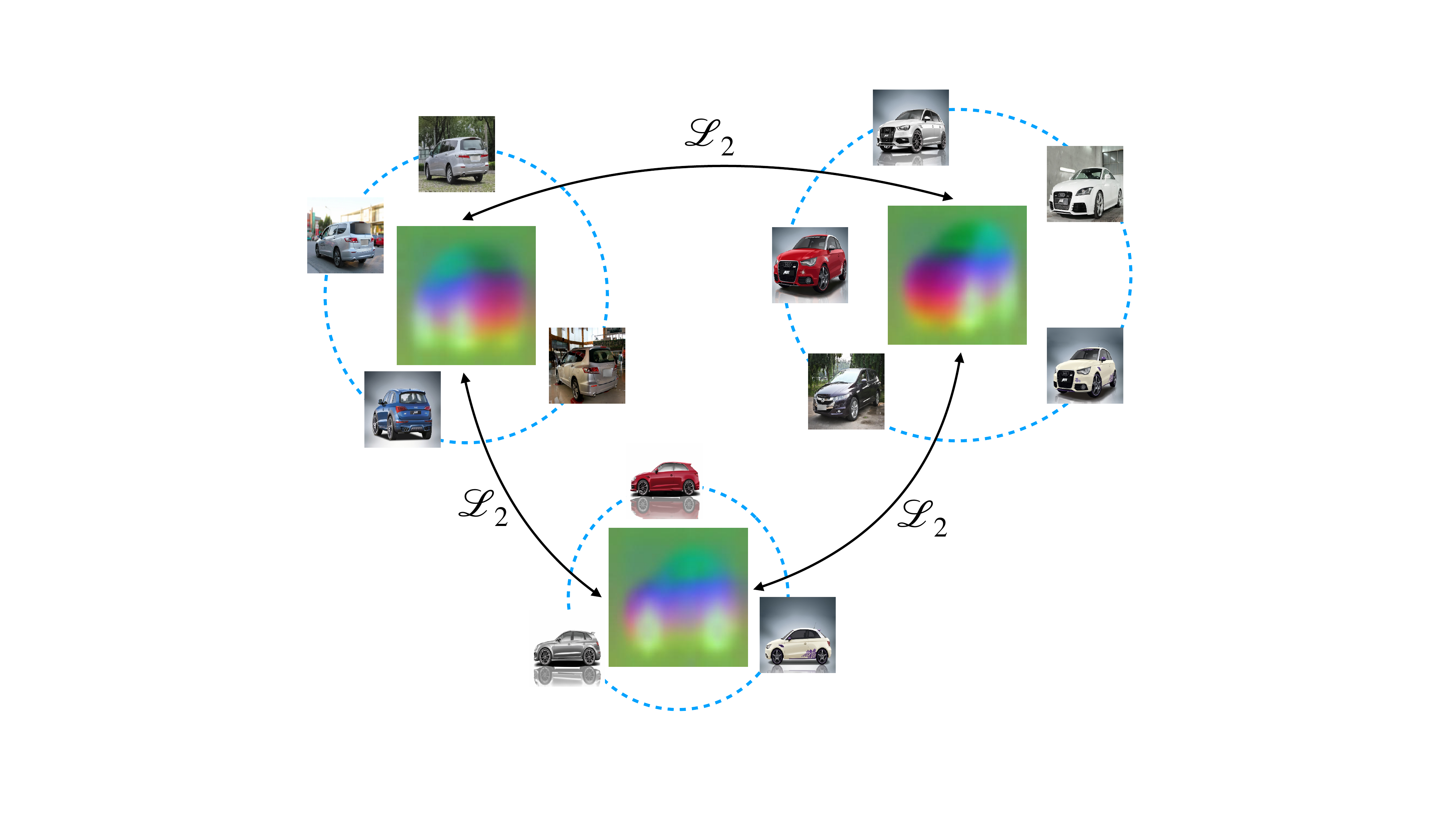}
        \caption{Pose-Centric Clustering}
        \label{fig:clustering}
    \end{subfigure} &
    \hfill
    \begin{subfigure}[a]{0.63\linewidth}
        \centering
        \includegraphics[width=\linewidth, trim=6cm 6.3cm 6cm 6.3cm, clip]{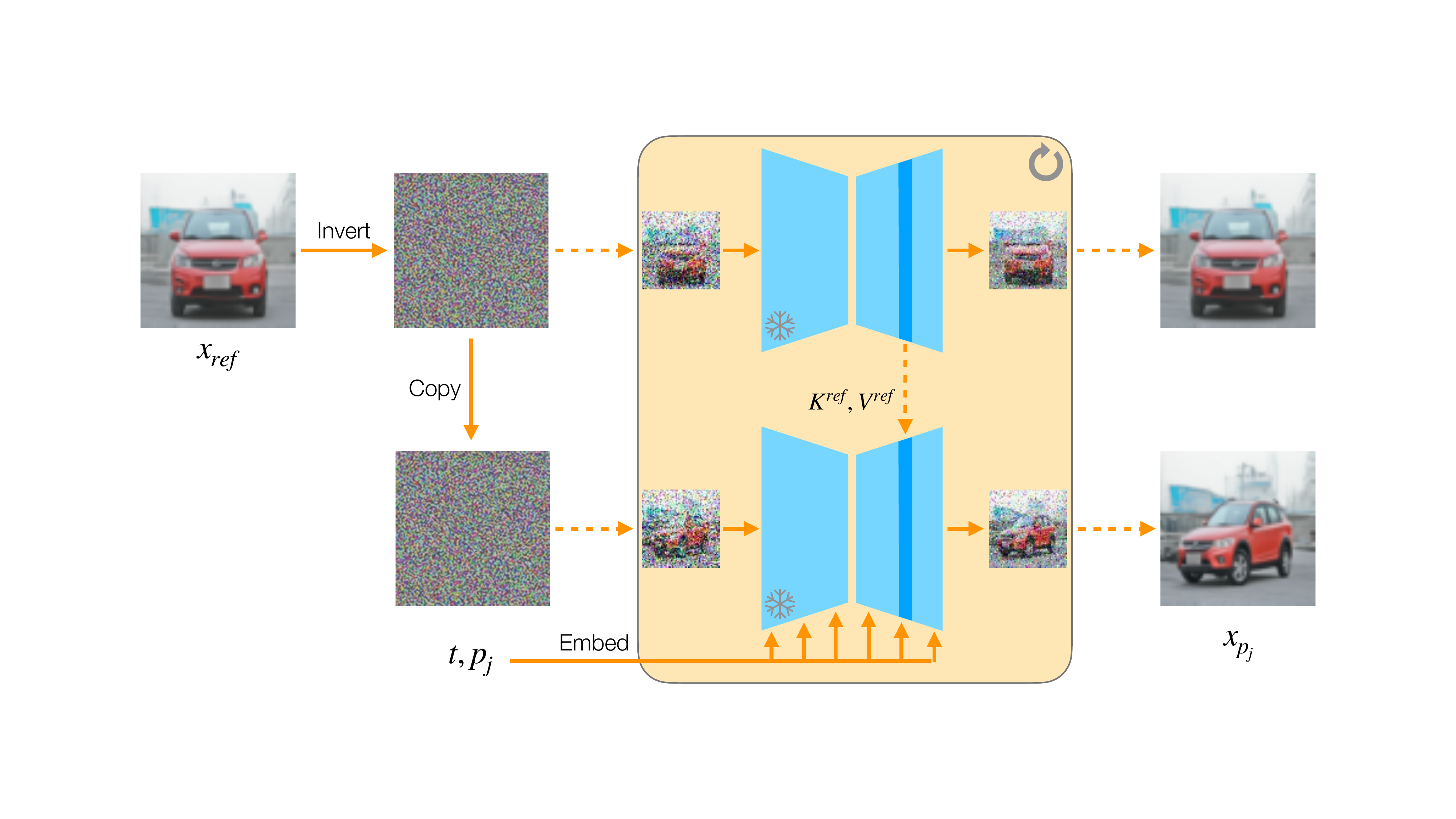}
        \caption{Inference with Cross-Frame Attention}
        \label{fig:cfa}
    \end{subfigure}
    \end{tabular}
    \caption{The illustration of the proposed pipeline for unsupervised novel view synthesis. (a) Pose-centric clustering based on ${\cal L}_2$ distance between first 3 principal components of DINOv2 spatial features after centering and rescaling. The average feature map is shown in the center of each cluster, upsampled for illustration purposes. (b) The inference with \methodName. The keys and values in the self-attention layers of the denoising backbone of the pose-conditioned diffusion are passed from the reference pose to generate the image of the object in the reference frame in pose $p_j$.}
    \label{fig:pipeline}
\end{figure*}

\subsection{Pose-Centric Clustering via Feature Visibility}\label{sec:clustering}

So far we assumed that the set of possible poses ${\cal P}$ is given. However, in the absence of such annotations one needs to discover ${\cal P}$ from ${\cal D}$ in an unsupervised way. To do so, we start by noticing that the pose of an object is defined by the set of visible semantic features and their locations in the image. For instance, when a car rotates, different parts of it (lights, wheels, doors, etc.) become visible at different locations, while other parts remain occluded. At the same time, for two different cars in the same pose after centering and rescaling certain parts appear in roughly similar locations. Thus, with respect to some metric that is based on comparing semantic local features of the objects after centering and rescaling the images, ${\cal D}$ splits into clusters that approximately correspond to different poses in the dataset.

As the feature extractor we adopt DINOv2~\cite{dinov2}, as its features were shown to capture semantics of object parts. As suggested in the original paper, we start by fitting PCA~\cite{pca} in the space of DINOv2 spatial features. Selecting the tokens that project to the positive axis of the first principal component allows us to roughly segment the object of interest in the image. Using the segmentation masks we center and rescale the images so that the objects span a shared predefined bounding box. We continue by fitting another PCA, this time only on the masked features (i.e. features that belong to the objects). We then project the original features to the first 3 principal components of the second PCA. This way we minimize the information about the texture in the features, while retaining the semantics of object parts (see Figure~\ref{fig:clustering}). The resulting tensors are compared in euclidean distance to cluster the dataset via K-Means clustering~\cite{Lloyd1982LeastSQ}. Figure~\ref{fig:clustering} showcases sample clusters after applying the described procedure. Since running PCA on the whole dataset is time-consuming, we opt for Incremental PCA~\cite{Ross2008IncrementalLF} that works at the batch-level and refines the approximate global PCA at each iteration. 

\subsection{Pose-Conditioned Diffusion}\label{sec:diffusion}

Given the pose-cluster assignments obtained with the procedure described in~\ref{sec:clustering}, we train a diffusion that generates images conditioned on those pose labels. More precisely, we train DDPM~\cite{Ho2020DenoisingDP} that learns to predict the noise added to the clean image by minimizing the following objective:

\begin{align}
    {\cal L}_{\text{DDPM}}(\theta) &=
    \sum_{t = 1}^{T}\mathbb{E}_{x_0, \varepsilon} \| \varepsilon^{(t)}_\theta(x_t; p) - \varepsilon\|^2, \\
    x_t &= \sqrt{\alpha_t} x_0 + \sqrt{1 - \alpha_t} \varepsilon, 
\end{align}

where the expectation with respect to $x_0$ is taken over the dataset ${\cal D}$, $\varepsilon$ is sampled from the standard normal distribution, and $p \in {\cal P}$ is the pose cluster assignment of $x_0$. $\{\alpha_t\}_{t = 1}^T$ is a sequence of decreasing positive real numbers defined such that the distribution of $x_t$ is approximately standard normal as $t \rightarrow T$ with a sufficiently large $T$.

At inference, we leverage deterministic DDIM sampler~\cite{song2020denoising} for efficiency. DDIM starts by sampling $x_T$ from ${\cal N}(0, 1)$ and iteratively denoises it by directly predicting $x_0$ from $x_t$ at each iteration and using it to predict $x_{t - 1}$ via the following formula:

\begin{equation}\label{eq:ddim}
\begin{split}
    x_{t - 1} = &\sqrt{\alpha_{t - 1}} \left(\frac{x_t - \sqrt{1 - \alpha_t} \varepsilon_\theta^{(t)}(x_t; p_j)}{\sqrt{\alpha_t}}\right) +\\ &\sqrt{1 - \alpha_{t - 1}} \varepsilon_\theta^{(t)}(x_t; p_j).
\end{split}
\end{equation}

A typical choice for $\varepsilon_\theta^{(t)}$ is the U-Net architecture~\cite{Ronneberger2015UNetCN} with group normalization~\cite{Wu2018GroupN} and built-in self-attention layers~\cite{Vaswani2017AttentionIA}. We condition $\varepsilon_\theta^{(t)}$ on the desired pose label $p_j$ by first embedding it to some vector space and then using this embedding to normalize the features in the residual blocks of the U-Net, similar to the AdaIN framework~\cite{Huang2017ArbitraryST}.

\subsection{Multi-View Generation}

While the conditioning on $p_j$ enables the diffusion introduced in Section~\ref{sec:diffusion} to generate images of objects in diverse and controlled poses, for novel view synthesis one needs to make sure that the generated images capture the same object. This is not guaranteed even when the images are generated starting from the same noise (see top row of Figure~\ref{fig:ablations}). Therefore, we propose to utilize an inference scheme similar to those used for video generation to ensure zero-shot temporal consistency of pretrained text-to-image diffusion models~\cite{text2video-zero}.

\noindent\textbf{Cross-attention is all you need.} It has been shown recently that the self-attention layers in the U-Net backbone of pretrained diffusion models capture structure (queries and keys) and appearance (values) of the objects in the image. Moreover, transferring keys and values from one image to another results in appearance transfer, while the global structure is preserved~\cite{Alaluf2023CrossImageAF}. This is achieved with the so called cross-frame attention procedure. Normally the self-attention layers take feature maps $f \in \mathbb{R}^{c\times h\times w}$ from the previous blocks of the network, linearly project them and reshape to obtain $Q, K$ and $V \in \mathbb{R}^{n \times d}$, where $n = h \times w$, and compute the output using the following formula:

\begin{align}
    \text{Self-Attention}(Q, K, V) = \text{Softmax}\left(\frac{Q K^\top}{\sqrt{d}}\right) V.
\end{align}

In the cross-frame attention the self-attention layers are modified to allow attending to the keys from the reference image only. More precisely, let $Q^{p_j}$ denote the queries calculated in one of the self-attention layers during the generation process with the pose label $p_j$ and $K^{\text{ref}}, V^{\text{ref}}$ be the corresponding key and values for the reference image. Then the output of the cross-frame attention (CFA) is calculated as:

\begin{align}\label{eq:cfa}
    \text{CFA}(Q^{p_j}, K^{\text{ref}}, V^{\text{ref}}) = \text{Softmax}\left(\frac{Q^{p_j} (K^{\text{ref}})^\top}{\sqrt{d}}\right) V^{\text{ref}}.
\end{align}

By using this technique we ensure that the global structure (i.e. the pose) of the object is aligned with the condition $p_j$, while the appearance, the local stucture and the identity are carried over from the reference image. This allows to generate valid novel views.

\noindent\textbf{Hard-attention guidance.} 
Moreover, we incorporate classifier-free guidance~\cite{Ho2022ClassifierFreeDG} into the denoising process. At each iteration, we replace $\varepsilon_\theta^{(t)}(x_t; p_j)$ in equation~\ref{eq:ddim} with $\varepsilon^{\text{g}}$, calculated as follows:

\begin{align}\label{eq:cfg}
    \varepsilon^{\text{g}}_t = (1 - \gamma) \cdot \varepsilon^{\text{hard}}_t + \gamma \cdot \varepsilon^{\text{soft}}_t,
\end{align}

Here, $\varepsilon^{\text{soft}}_t$ denotes the output of the denoising network using cross-frame attention, and $\varepsilon^{\text{hard}}_t$ follows the same procedure as $\varepsilon^{\text{soft}}_t$ but with Argmax instead of Softmax in equation~\ref{eq:cfa}. We call this procedure Hard-Attention Guidance, or HAG. It is worth noting that, in contrast to conventional classifier-free guidance, equation~\ref{eq:cfg}, with $\gamma \geq 0$, defines a negative guidance. In our case $\gamma = 1$ means no HAG is applied. This negative guidance directs the generation process away from situations where similar parts of the object hardly attend to different parts in the reference image, resulting in inconsistent artifacts in the generated sequences. This technique not only improves the quality of generated images but also increases consistency across views, as demonstrated in our experiments (see Figure~\ref{fig:ablations}).

\noindent\textbf{Single image to multi-view.}
To generate new views of a given image, the initial step involves inverting the diffusion process to extract the noise responsible for generating that particular sample. For this, we exploit the deterministic nature of the DDIM~\cite{song2020denoising} sampler, which allows a straightforward inversion. Once the latent noise is identified, it is used to generate other views with poses that are different from the one in the reference image. Throughout this novel view generation process, keys and values are always transferred from the reference image. This approach guarantees that the generated views maintain coherence with the input image.

\section{Experiments}
\label{sec:experiments}

In this section, we present implementation details and results.
For more comprehensive ablation studies as well as for more visual results, we refer the reader to the supplementary material.

\begin{table}[t]
\centering
\begin{tabular}{c l r }
\toprule
Dataset & Method & FID $\downarrow$ \\ \hline
\multirow{5}{*}{CompCars~\cite{compcars}} & Ours - no MV &  6.081 \\
& Ours - MV, w/ bcg, w/o HAG &  9.320 \\
& Ours - MV, w/ bcg, w/ HAG & 9.079  \\
& Monnier \etal~\cite{monnier2022unicorn} - w/o bcg & 115.560 \\ 
& Ours - MV, w/o bcg & 34.796 \\ \hline
\multirow{3}{*}{SD Armchairs} & Ours - no MV & 5.744  \\
& Ours - MV, w/o HAG & 7.870 \\
& Ours - MV, w/ HAG &  11.204 \\ \hline
\end{tabular}
\caption{The FID scores for CompCars and SD Armchairs for various versions of \methodName (ours) and Monnier \etal~\cite{monnier2022unicorn} method. MV refers to multi-view and bcg is an abbreviation for background.} \label{table:FID}
\end{table}


\subsection{Miscellaneous}

\noindent\textbf{Datasets.} To assess the performance of our method we carry out experiments on the \textbf{CompCars}~\cite{compcars} dataset that contains up to $100K$ images of cars captured from various poses ranging in 360 degrees. Lacking large and diverse datasets of single-category images, in order to assess the robustness of our method to textures and shapes other than cars, we propose to leverage Stable Diffusion~\cite{rombach2021highresolution} (SD) to generate pseudo-real datasets of that nature. More precisely, we condition SD on prompts of the kind \textit{``A high-quality photo of \textlangle category\textrangle, \textlangle view-point\textrangle''}, where the view-point can be \textit{``front view''}, \textit{``side view''} and \textit{``back view''}. Additionally, we specify negatives prompts as \textit{``Low-quality, close-up, cropped, [view-point]''}, where the optional view-point field is set to \textit{``front view''} for the positive prompts featuring back and side views. In this paper we consider the \textit{``single massive armchair''} category, while other categories are to be explored in the future work. We generate 100k images of armchairs that form the \textbf{SD~Armchairs} dataset. We will release this dataset upon publication.

\noindent\textbf{Implementation Details.}
For both datasets, we train our models on images of size $64\times64$. During training, we adhere to the hyperparameters outlined in DDIM~\cite{song2020denoising}, i.e., setting the learning rate to $2\times10^{-4}$ and utilizing the Adam optimizer~\cite{kingma2017adam}. 
During inference, we set the guidance strength to $\gamma=1.5$ for all datasets and employ $50$ denoising steps. 
Note that we apply our cross-attention framework solely in the bottleneck and the decoder of the U-Net, as we found that using it in the encoder does not have an effect.
In the case of the CompCars dataset, we opt for a cluster count of 25, while for armchairs, we choose 10 clusters.
In particular, we observed that in both cases, some clusters shared the same pose. 
It is worth noting that we found a trade-off between the number of clusters and the pureness of the poses inside the cluster. More specifically, when we attempted to reduce the number of clusters, we found that images within clusters started to exhibit more diverse (noisy) poses/orientations.


\noindent\textbf{Baseline.}
To the best of our knowledge, our method is the first approach that proposes novel view synthesis without the reliance on any form of supervision. In our evaluation, we compare with the most closely related method, specifically that of Monnier et al.~\cite{monnier2022unicorn}, which reconstructs objects in an unsupervised manner. For the comparisons on the car dataset, we utilize their official pretrained models\footnote{\href{https://github.com/monniert/unicorn}{https://github.com/monniert/unicorn}}. 
However, in the case of the armchair dataset, we encountered difficulties in training their model. We hypothesize that hyperparameters play a crucial role in the convergence of their model and experiments with various random seeds are required to achieve optimal results.

\subsection{Ablations}

\begin{figure}
    \centering
    \footnotesize
    \begin{tblr}{
      rows = {belowsep=0pt},
      column{5,6,7} = {wd=0.19\linewidth, halign=c, colsep=1pt},
      column{4} = {wd=0.19\linewidth, halign=c, leftsep=1pt, rightsep=1pt},
      column{1,2} = {wd=0.1cm, halign=c, valign=m, colsep=6pt},
      column{3} = {wd=0.2cm, halign=c, valign=m, leftsep=6pt, rightsep=6pt},
      vline{4} = {1pt},
      vline{5} = {1pt},
      cell{2-8}{1-3} = {}
    }
         \rotatebox{90}{\small same noise} & \rotatebox{90}{\small CFA} & \rotatebox{90}{\large $\gamma$ \small HAG} & \rotatebox{0}{reference} & \SetCell[c=3]{c} generated novel views \\
         \hline
         \checkmark & & 1.0 &
         \raisebox{-.5\height}{\includegraphics[width=\linewidth]{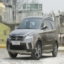}} &  
         \raisebox{-.5\height}{\includegraphics[width=\linewidth]{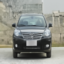}} & 
         \raisebox{-.5\height}{\includegraphics[width=\linewidth]{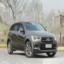}} & 
         \raisebox{-.5\height}{\includegraphics[width=\linewidth]{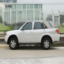}} \\
          & \checkmark & 1.0 &
         \raisebox{-.5\height}{\includegraphics[width=\linewidth]{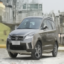}} &  
         \raisebox{-.5\height}{\includegraphics[width=\linewidth]{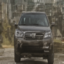}} & 
         \raisebox{-.5\height}{\includegraphics[width=\linewidth]{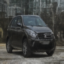}} & 
         \raisebox{-.5\height}{\includegraphics[width=\linewidth]{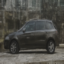}} \\
         \checkmark & \checkmark & 0.5 & 
         \raisebox{-.5\height}{\includegraphics[width=\linewidth]{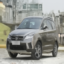}} &  
         \raisebox{-.5\height}{\includegraphics[width=\linewidth]{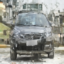}} & 
         \raisebox{-.5\height}{\includegraphics[width=\linewidth]{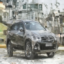}} & 
         \raisebox{-.5\height}{\includegraphics[width=\linewidth]{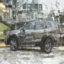}} \\
         \checkmark & \checkmark & 1.0 &
         \raisebox{-.5\height}{\includegraphics[width=\linewidth]{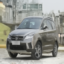}} &  
         \raisebox{-.5\height}{\includegraphics[width=\linewidth]{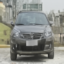}} & 
         \raisebox{-.5\height}{\includegraphics[width=\linewidth]{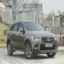}} & 
         \raisebox{-.5\height}{\includegraphics[width=\linewidth]{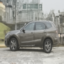}} \\
         \checkmark & \checkmark & 1.5 & 
         \raisebox{-.5\height}{\includegraphics[width=\linewidth]{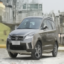}} &  
         \raisebox{-.5\height}{\includegraphics[width=\linewidth]{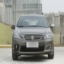}} & 
         \raisebox{-.5\height}{\includegraphics[width=\linewidth]{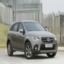}} & 
         \raisebox{-.5\height}{\includegraphics[width=\linewidth]{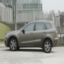}} \\
         \checkmark & \checkmark & 2.0 & 
         \raisebox{-.5\height}{\includegraphics[width=\linewidth]{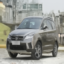}} &  
         \raisebox{-.5\height}{\includegraphics[width=\linewidth]{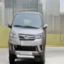}} & 
         \raisebox{-.5\height}{\includegraphics[width=\linewidth]{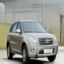}} & 
         \raisebox{-.5\height}{\includegraphics[width=\linewidth]{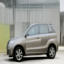}} \\
         \checkmark & \checkmark & 2.5 & 
         \raisebox{-.5\height}{\includegraphics[width=\linewidth]{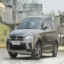}} &  
         \raisebox{-.5\height}{\includegraphics[width=\linewidth]{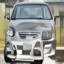}} & 
         \raisebox{-.5\height}{\includegraphics[width=\linewidth]{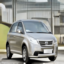}} & 
         \raisebox{-.5\height}{\includegraphics[width=\linewidth]{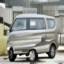}}
    \end{tblr}
    \caption{Ablations of the usage of the same  initial noise at generation, cross-frame attention (CFA) and the strength of hard-attention guidance ($\gamma$ HAG). The combination of CFA and generation from the same noise ensures consistency across view, while the usage of HAG imposes smoothness in the generated views and removes inconsistent artifacts.}
    \label{fig:ablations}
\end{figure}

In this section we ablate some of our design choices. We start by emphasizing the importance of using CFA and HAG for obtaining consistent and smooth novel views (see Figure~\ref{fig:ablations}). Increasing the strength of HAG smoothens the results. However, too large values of $\gamma$ result in unrealistic images. Besides this, we also noticed that the global statistics of the images are encoded in the initial noise that is used to generate the views. Therefore, starting the generation from the same noise improves consistency.

\begin{figure*}
\begin{center}
    \centering
    \captionsetup{type=figure}
        \begin{tabular}{@{}c
        @{\hspace{0.5mm}}c@{\hspace{0.5mm}}c@{\hspace{0.5mm}}c@{\hspace{0.5mm}}c@{\hspace{0.5mm}}c@{\hspace{0.5mm}}c@{\hspace{0.5mm}}c@{\hspace{0.5mm}}c@{}} 
            Original & Inverted  & \multicolumn{6}{c}{Generated Novel Views}  \\
            {\includegraphics[width=0.124\linewidth]{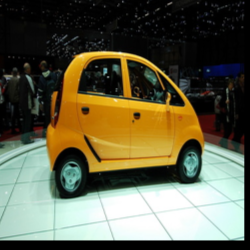}} &
            {\includegraphics[width=0.124\linewidth]{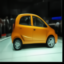}} & {\includegraphics[width=0.124\linewidth]{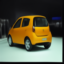}} & 
            {\includegraphics[width=0.124\linewidth]{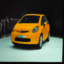}} & 
            {\includegraphics[width=0.124\linewidth]{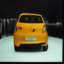}} &
            {\includegraphics[width=0.124\linewidth]{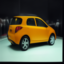}} & 
            {\includegraphics[width=0.124\linewidth]{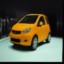}} & 
            {\includegraphics[width=0.124\linewidth]{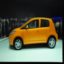}} \\

            {\includegraphics[width=0.124\linewidth]{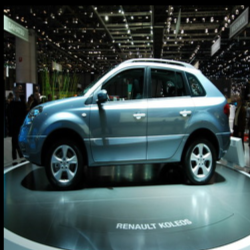}} &
            {\includegraphics[width=0.124\linewidth]{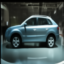}} & {\includegraphics[width=0.124\linewidth]{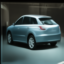}} & 
            {\includegraphics[width=0.124\linewidth]{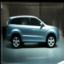}} & 
            {\includegraphics[width=0.124\linewidth]{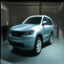}} &
            {\includegraphics[width=0.124\linewidth]{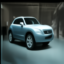}} & 
            {\includegraphics[width=0.124\linewidth]{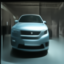}} & 
            {\includegraphics[width=0.124\linewidth]{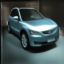}} \\

            {\includegraphics[width=0.124\linewidth]{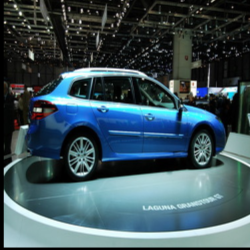}} &
            {\includegraphics[width=0.124\linewidth]{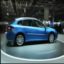}} & {\includegraphics[width=0.124\linewidth]{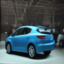}} & 
            {\includegraphics[width=0.124\linewidth]{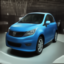}} & 
            {\includegraphics[width=0.124\linewidth]{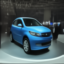}} &
            {\includegraphics[width=0.124\linewidth]{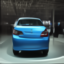}} & 
            {\includegraphics[width=0.124\linewidth]{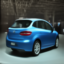}} & 
            {\includegraphics[width=0.124\linewidth]{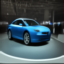}} \\
            
        \end{tabular}
    \captionof{figure}{
    Additional examples of novel views generated by \methodName on the EPFL Dataset.\label{fig:epfl_novel}}
\end{center}
\end{figure*}

\begin{figure}
\begin{center}
    \centering
    \captionsetup{type=figure}
        \begin{tabular}{@{}r@{\hspace{1.5mm}}c@{\hspace{0.5mm}}c@{\hspace{0.5mm}}c@{\hspace{0.5mm}}c@{\hspace{0.5mm}}c@{\hspace{0.5mm}}c@{\hspace{0.5mm}}c@{\hspace{0.5mm}}c@{}} 
            Reference & \multicolumn{4}{c}{Generated Novel Views}  \\
            {\includegraphics[width=0.2\linewidth]{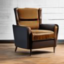}} &
            {\includegraphics[width=0.2\linewidth]{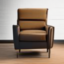}} & {\includegraphics[width=0.2\linewidth]{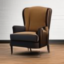}} & 
            {\includegraphics[width=0.2\linewidth]{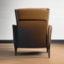}} & 
            {\includegraphics[width=0.2\linewidth]{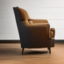}} &  \\

            {\includegraphics[width=0.2\linewidth]{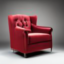}} &
            {\includegraphics[width=0.2\linewidth]{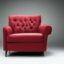}} & {\includegraphics[width=0.2\linewidth]{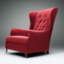}} & 
            {\includegraphics[width=0.2\linewidth]{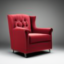}} & 
            {\includegraphics[width=0.2\linewidth]{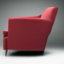}} &  \\

            {\includegraphics[width=0.2\linewidth]{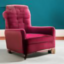}} &
            {\includegraphics[width=0.2\linewidth]{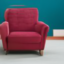}} & {\includegraphics[width=0.2\linewidth]{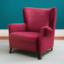}} & 
            {\includegraphics[width=0.2\linewidth]{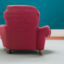}} & 
            {\includegraphics[width=0.2\linewidth]{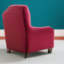}} &  \\

        \end{tabular}
    \captionof{figure}{
    Aditional examples of novel views generated by \methodName on the SD Armchairs dataset.\label{fig:armchairs_generated}}
\end{center}
\end{figure}

\subsection{Results}

We present the FID scores~\cite{fid} in Table~\ref{table:FID} for both datasets. The FID no-MV indicates that the images are generated without cross-attention, \ie, in a standard way, and they have the lowest FID score in both datasets among other scores.
We also computed the FID  for our generated multiview dataset. More specifically for each reference image (noise) we generated 3 novel views. For fairness, we randomly selected the pose reference for each instance (noise). 
We generated the multiview dataset for both w and w/o guidance. 
For the CompCars dataset we observe the version w/ guidance having lower FID, while for the SD Armchairs dataset the results are the opposite. We hypothesize  this to be the result of over-smoothing caused by HAG, since the textures in the ground truth data are already quite monotone.

\begin{figure}[h!]
\begin{center}
    \centering
    \captionsetup{type=figure}
        \begin{tabular}{@{}r@{\hspace{0.5mm}}c@{\hspace{0.5mm}}c@{\hspace{0.5mm}}c@{\hspace{0.5mm}}c@{\hspace{0.5mm}}c@{\hspace{0.5mm}}c@{\hspace{0.5mm}}c@{}} 
            {\includegraphics[width=.140\linewidth]{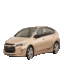}} & 
            {\includegraphics[width=.140\linewidth]{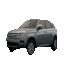}} & 
            {\includegraphics[width=.140\linewidth]{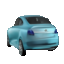}} & 
            {\includegraphics[width=.140\linewidth]{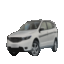}} &
            {\includegraphics[width=.140\linewidth]{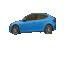}} &
            {\includegraphics[width=.140\linewidth]{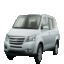}} \\

            {\includegraphics[width=.140\linewidth]{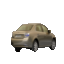}} & 
            {\includegraphics[width=.140\linewidth]{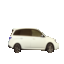}} & 
            {\includegraphics[width=.140\linewidth]{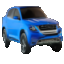}} & 
            {\includegraphics[width=.140\linewidth]{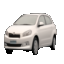}} &
            {\includegraphics[width=.140\linewidth]{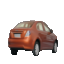}} &
            {\includegraphics[width=.140\linewidth]{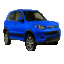}} \\

            {\includegraphics[width=.140\linewidth]{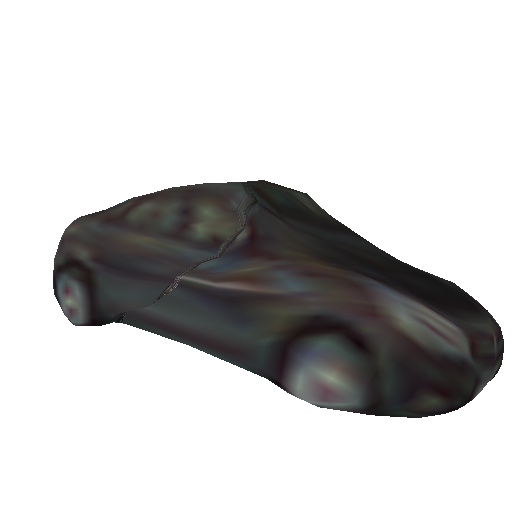}} & 
            {\includegraphics[width=.140\linewidth]{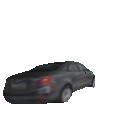}} & 
            {\includegraphics[width=.140\linewidth]{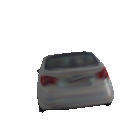}} & 
            {\includegraphics[width=.140\linewidth]{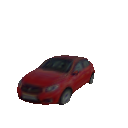}} &
            {\includegraphics[width=.140\linewidth]{figures/unicorn_images/rendered_image_000190.png}} &
            {\includegraphics[width=.140\linewidth]{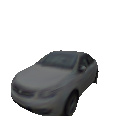}} \\
            
            {\includegraphics[width=.140\linewidth]{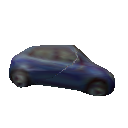}} & {\includegraphics[width=.140\linewidth]{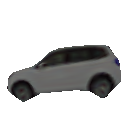}} & 
            {\includegraphics[width=.140\linewidth]{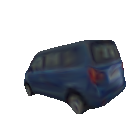}} & 
            {\includegraphics[width=.140\linewidth]{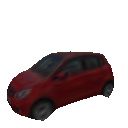}} &
            {\includegraphics[width=.140\linewidth]{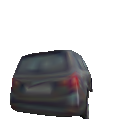}}&
            {\includegraphics[width=.140\linewidth]{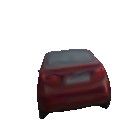}} \\
        
        \end{tabular}
    \captionof{figure}{
     Top 2 rows - Generated views from MIRAGE (ours). Bottom 2 row - Generated views from the reconstructed cars in the CompCars dataset using the method of Monnier \etal~\cite{monnier2022unicorn}.\label{fig:unicorn-images}}
\end{center}
\end{figure}

We also calculated the FID for Monnier \etal~\cite{monnier2022unicorn}. 
Since their method is not generative we took the CompCars dataset and reconstructed randomly selected 30K cars and then generated 4 novel views from 15K cars from this set. 
Since their generated  views have a white background we utilized SAM~\cite{sam_segmentor} to remove the background from the ground truth data. 
We also applied the same procedure to our generated dataset. From Table~\ref{table:FID}, we observe that the FID for Monnier \etal~\cite{monnier2022unicorn} is much higher compared to the our masked  version. This can also be observed visually in Figure~\ref{fig:unicorn-images}. 
This is expected as their method is not supposed to generate high quality textures because of the cross consistency loss between similar instances.

\begin{figure*}
\begin{center}
    \centering
    \captionsetup{type=figure}
        \begin{tabular}{@{}c@{\hspace{1.7mm}}c@{\hspace{0.7mm}}c@{\hspace{0.7mm}}c@{\hspace{0.7mm}}c@{\hspace{0.7mm}}c@{\hspace{0.7mm}}c@{\hspace{0.7mm}}c@{\hspace{0.7mm}}c@{}} 
            Reference & \multicolumn{6}{c}{Generated Novel Views}  \\
             {\includegraphics[width=0.130\linewidth]{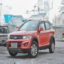}} & 
            {\includegraphics[width=0.130\linewidth]{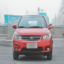}} & 
            {\includegraphics[width=0.130\linewidth]{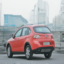}} &
            {\includegraphics[width=0.130\linewidth]{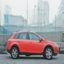}} & 
            {\includegraphics[width=0.130\linewidth]{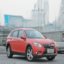}} & 
            {\includegraphics[width=0.130\linewidth]{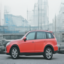}} & 
             {\includegraphics[width=0.130\linewidth]{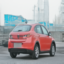}} 
            \\

             {\includegraphics[width=0.130\linewidth]{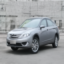}} & 
             {\includegraphics[width=0.130\linewidth]{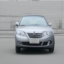}} & 
             {\includegraphics[width=0.130\linewidth]{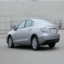}} &
             {\includegraphics[width=0.130\linewidth]{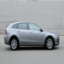}} & 
             {\includegraphics[width=0.130\linewidth]{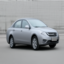}} & 
             {\includegraphics[width=0.130\linewidth]{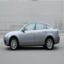}} 
            & {\includegraphics[width=0.130\linewidth]{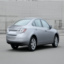}} 
            \\
            
            {\includegraphics[width=0.130\linewidth]{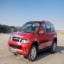}} & 
            {\includegraphics[width=0.130\linewidth]{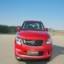}} & 
            {\includegraphics[width=0.130\linewidth]{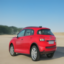}} &
            {\includegraphics[width=0.130\linewidth]{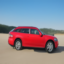}} & 
            {\includegraphics[width=0.130\linewidth]{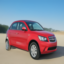}} & 
            {\includegraphics[width=0.130\linewidth]  {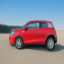}} &     {\includegraphics[width=0.130\linewidth]{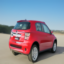}} \\
            
        \end{tabular}
    \captionof{figure}{
    Additional examples of novel views generated by \methodName on the CompCars dataset.\label{fig:compcars_generated}}
\end{center}
\end{figure*}

\begin{figure*}[ht]
\begin{center}
    \centering
    \captionsetup{type=figure}
        \begin{tabular}{@{}c@{\hspace{1.5mm}}c@{\hspace{0.5mm}}c@{\hspace{0.5mm}}c@{\hspace{0.5mm}}c@{\hspace{0.5mm}}c@{\hspace{0.5mm}}c@{\hspace{0.5mm}}c@{\hspace{0.5mm}}c@{\hspace{0.5mm}}c@{\hspace{0.5mm}}c@{\hspace{0.5mm}}c@{}}
            Input & \multicolumn{7}{c}{Novel Views} & \multicolumn{4}{c}{3D Meshes}  \\
            
            {\includegraphics[width=.08\linewidth]{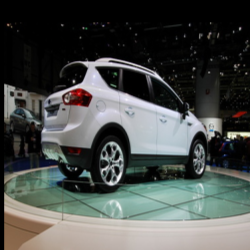}} & 
            {\includegraphics[width=.08\linewidth]{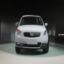}} & 
            {\includegraphics[width=.08\linewidth]{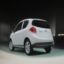}} & 
            {\includegraphics[width=.08\linewidth]{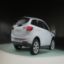}} &
            {\includegraphics[width=.08\linewidth]{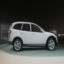}} & 
            {\includegraphics[width=.08\linewidth]{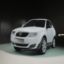}} & 
            {\includegraphics[width=.08\linewidth]{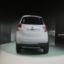}}& 
            {\includegraphics[width=.08\linewidth]{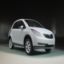}}& 
            {\includegraphics[width=.08\linewidth]{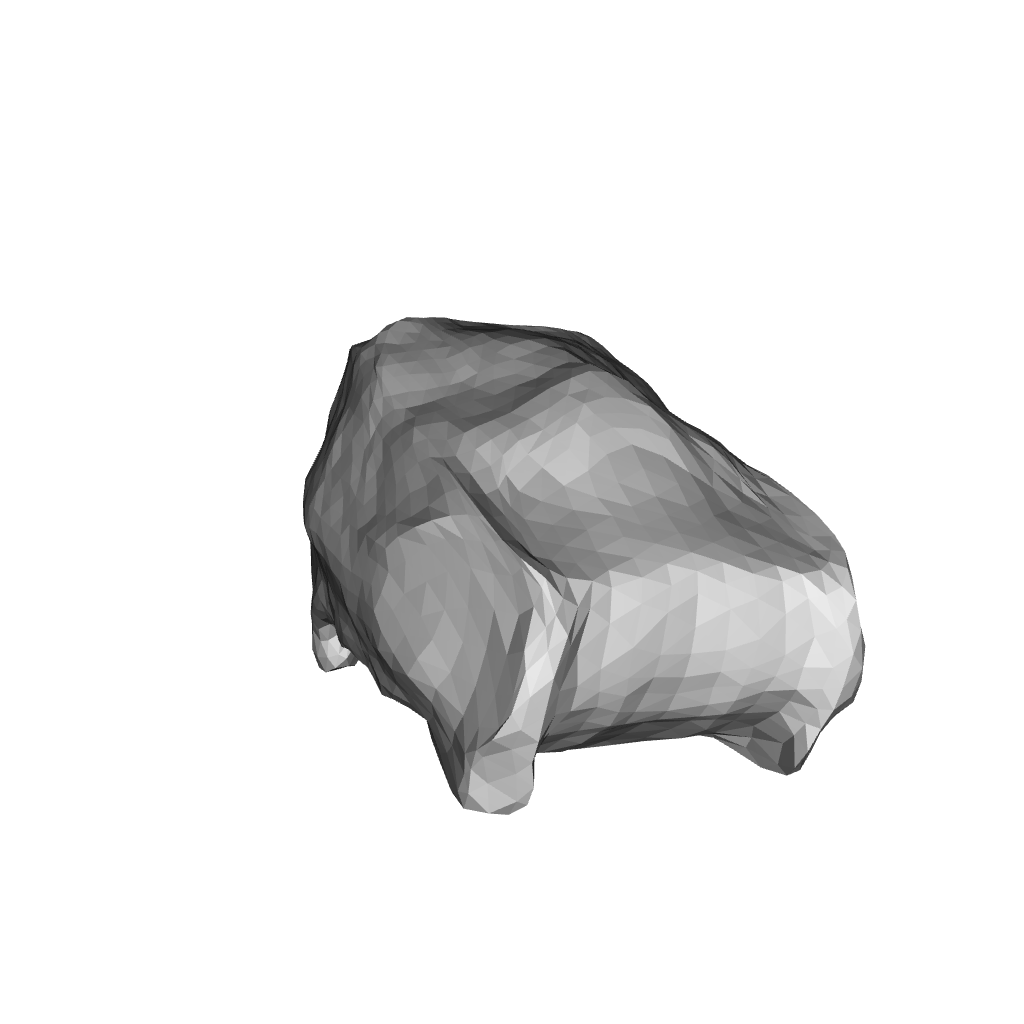}}& 
            {\includegraphics[width=.08\linewidth]{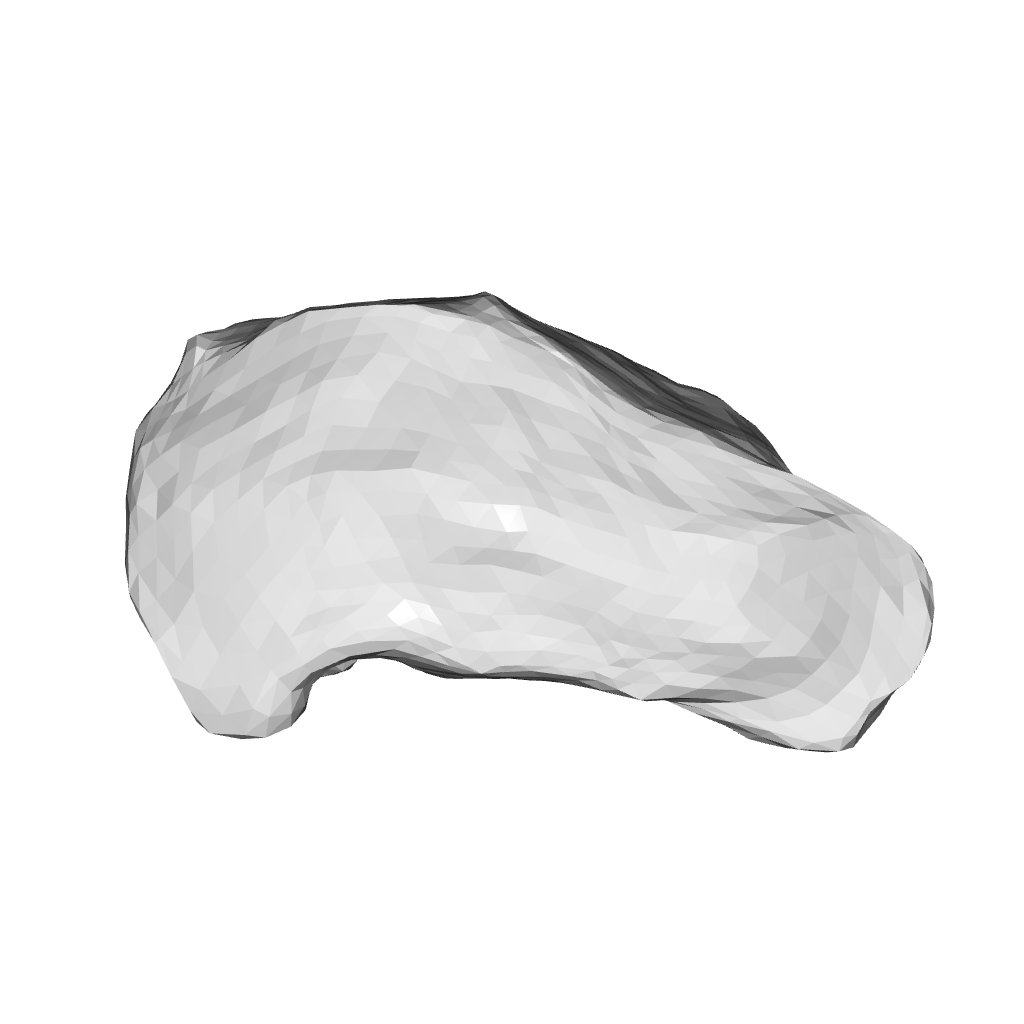}}& 
            {\includegraphics[width=.08\linewidth]{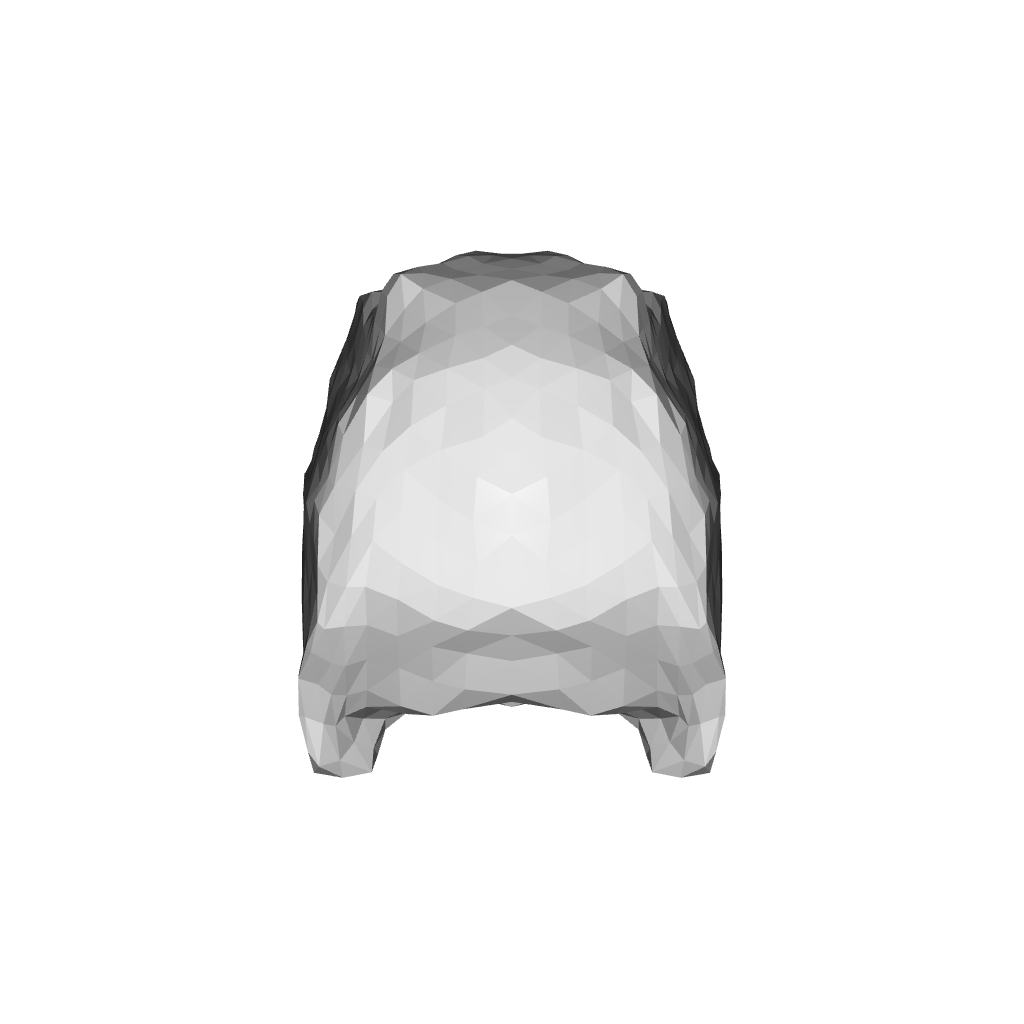}}& 
            {\includegraphics[width=.08\linewidth]{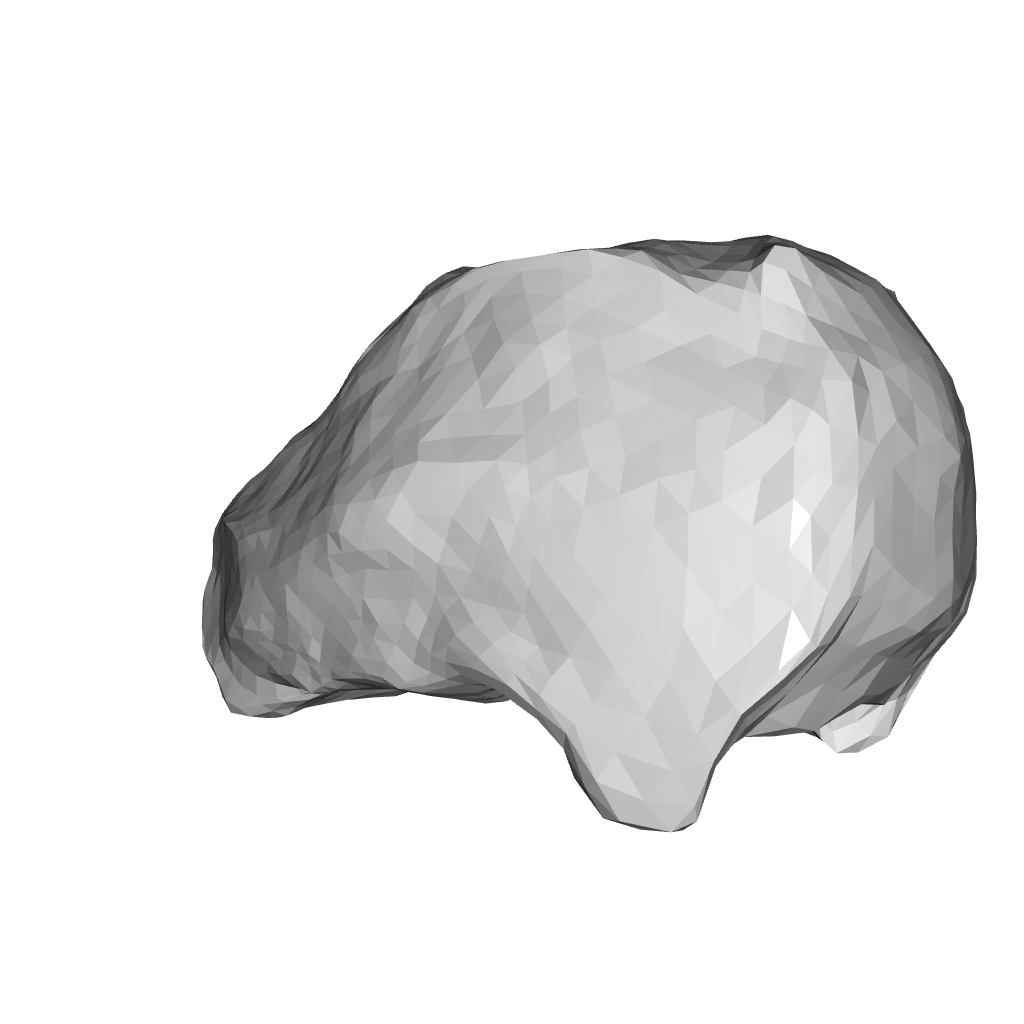}} \\
             & {\includegraphics[width=.08\linewidth]{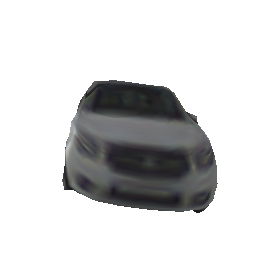}} & 
            {\includegraphics[width=.08\linewidth]{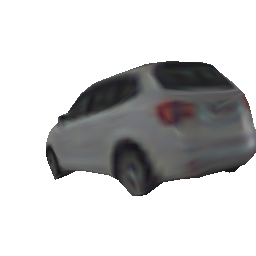}} & 
            {\includegraphics[width=.08\linewidth]{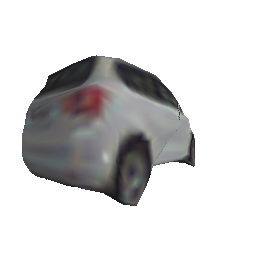}} &
            {\includegraphics[width=.08\linewidth]{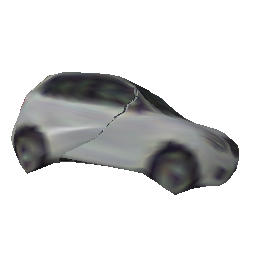}} & 
            {\includegraphics[width=.08\linewidth]{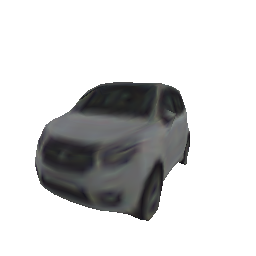}} & 
            {\includegraphics[width=.08\linewidth]{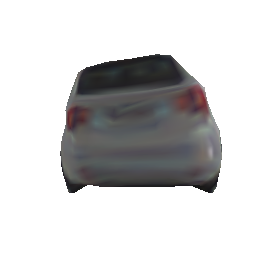}}& 
            {\includegraphics[width=.08\linewidth]{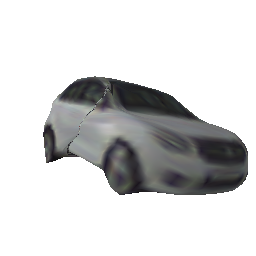}} & 
            {\includegraphics[width=.08\linewidth]{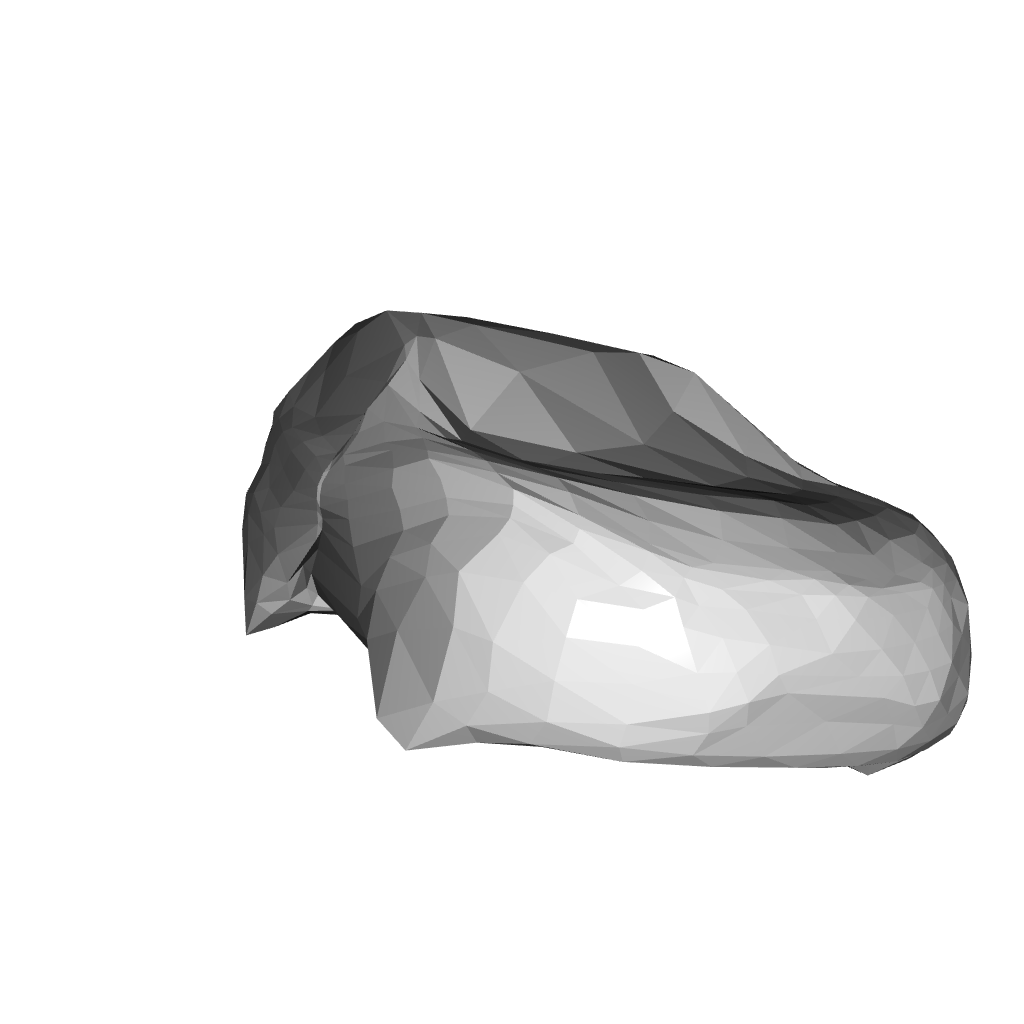}}& 
            {\includegraphics[width=.08\linewidth]{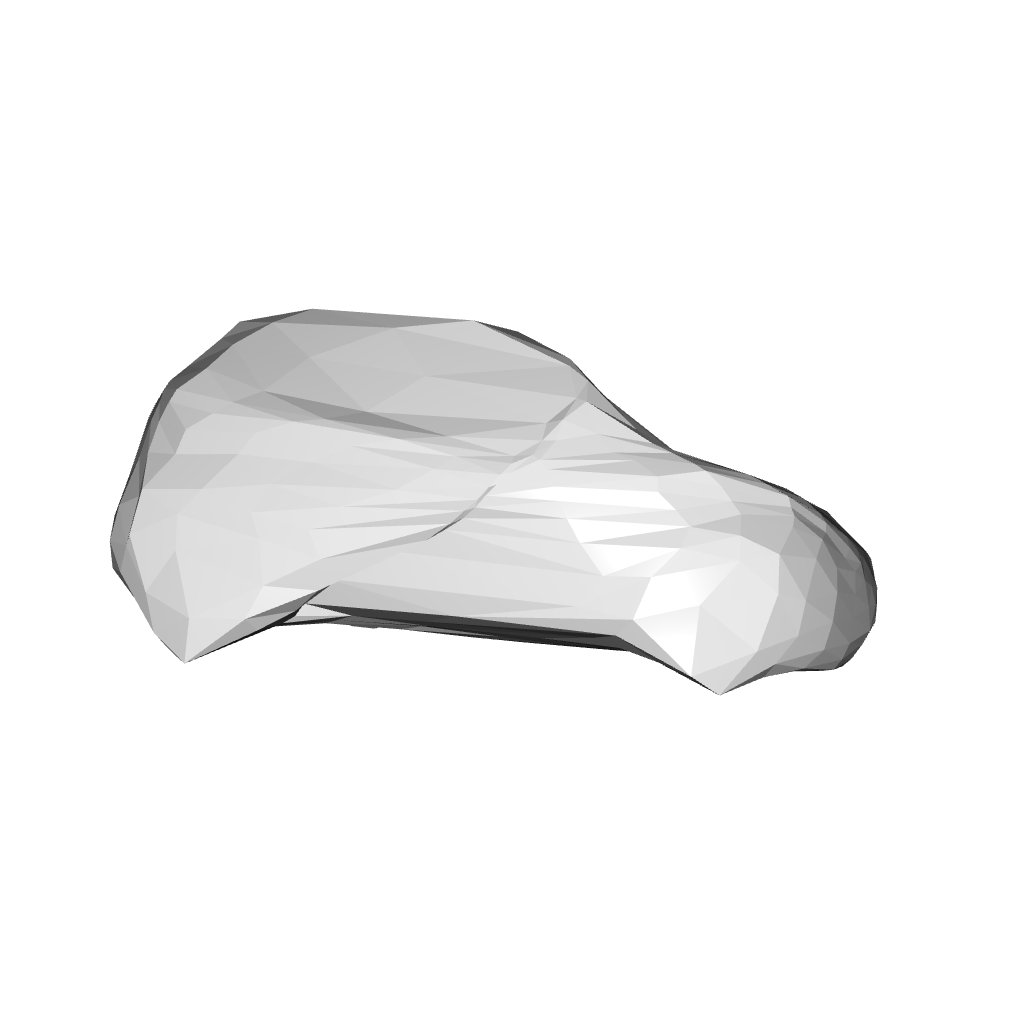}}& 
            {\includegraphics[width=.08\linewidth]{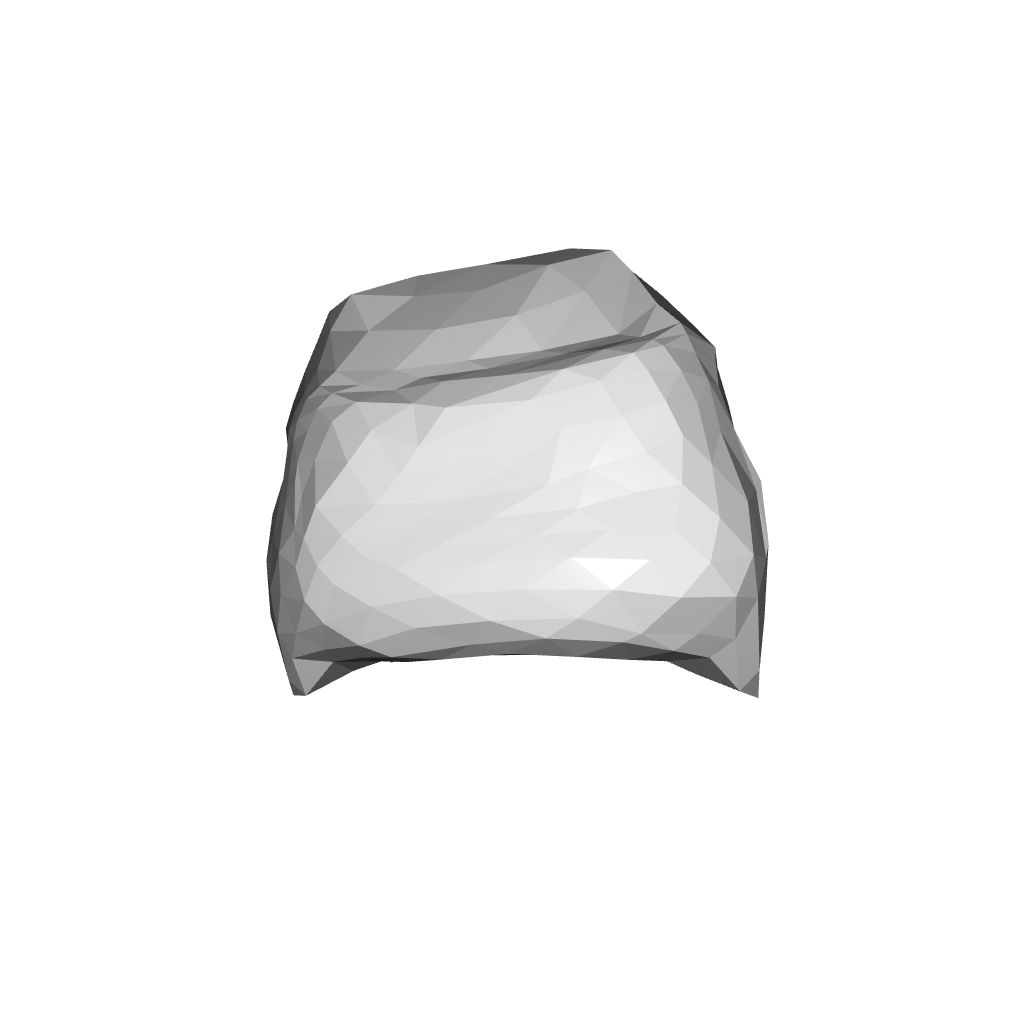}}& 
            {\includegraphics[width=.08\linewidth]{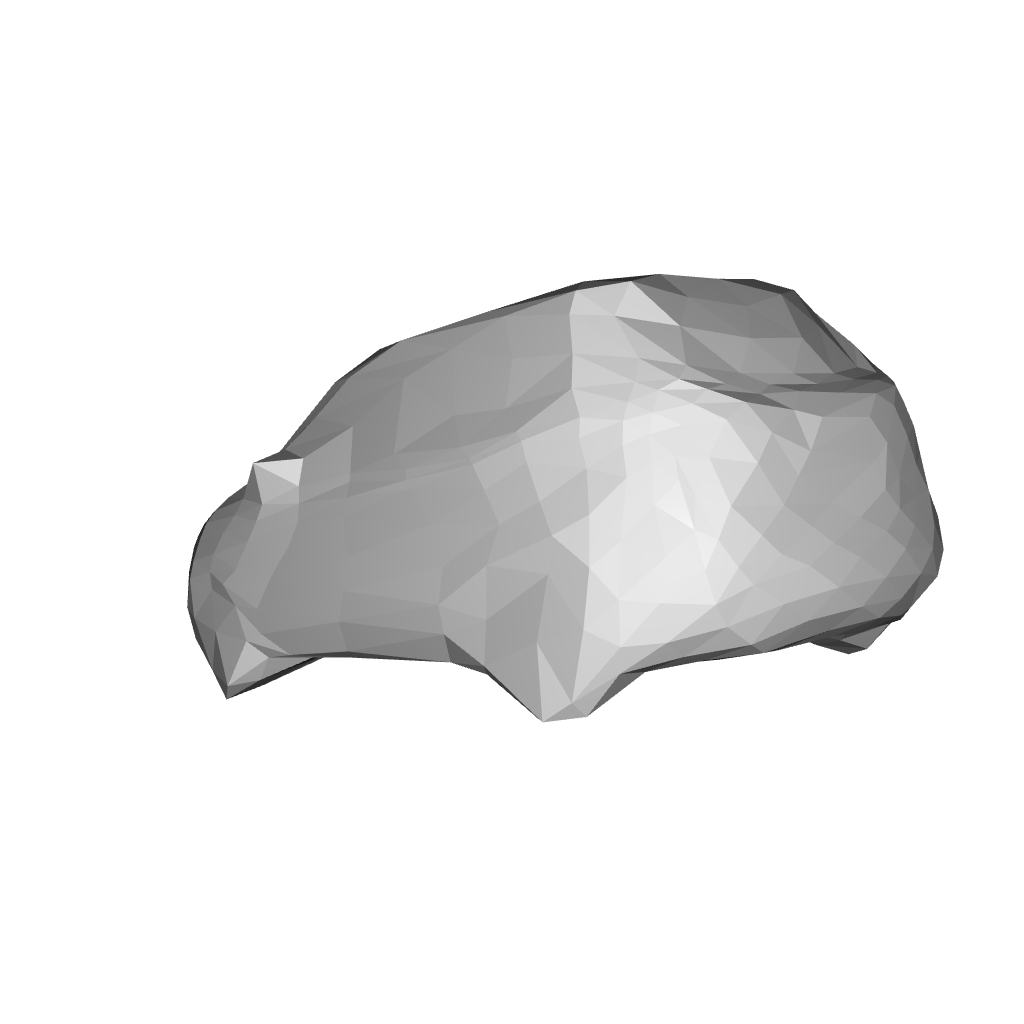}} \\

        \end{tabular}
    \captionof{figure}{
    Reconstructed 3D for a sample car in the EPFL dataset. Top row - \methodName (ours), bottom row - Monnier~\etal~\cite{monnier2022unicorn}. Notice that the latter does not fit backgrounds in the model, that is why there are no backgrounds in the novel views.\label{fig:3D_RECON}}
\end{center}
\end{figure*}

In Figures~\ref{fig:compcars_generated} and \ref{fig:armchairs_generated} we show visual results for both generated novel views for car and armchair datasets starting from the normal noise. 
In Figure~\ref{fig:epfl_novel} we show the novel views for some cars in the EPFL dataset~\cite{epfl_car_dataset}, which is out-of-distribution for the model trained on CompCars. For this, we first take the test image and then  find its pose label using the procedure explained in section~\ref{sec:clustering}. Once we do the inversion, we generate novel views as  described in section~\ref{sec:method}.

\noindent\textbf{3D Reconstruction}. In Figure~\ref{fig:3D_RECON} we showcase the obtained 3D model from novel views generated for the car from the EPFL dataset. Once we obtained novel views from our method we run NeRS~\cite{zhang2021ners} on a sparse set of novel views. Here we have to we initialize the camera poses manually. We use SAM~\cite{sam_segmentor} to obtain the masks since NeRS can work only with mask supervision. In the absence of camera poses it was difficult to assign initial poses to the generated images. For this reason,  we run only on a sparse set of generated images. In Figure~\ref{fig:3D_RECON}, we show the 3D-mesh obtained by running NeRS on our generated images. We observe our novel views to be good enough for obtaining explicit 3D-Mesh. 
Note that our generated views are more detailed in terms of texture than the ones of  Monnier~\etal~\cite{monnier2022unicorn}.




\section{Conclusion and Limitations}

This paper introduces \methodName, a novel pipeline designed for generating multi-view data without relying on manual annotations. Our approach outperforms prior methods in generating novel view sequences and exhibits robustness across various domains. However, a limitation is that it requires training on a dataset of single-category images. Developing an unsupervised approach for novel view synthesis on multi-category data is left for future work. We hypothesise that multi-category data can be hierarchically clustered, first to object classes (\eg, by comparing DINOv2 cls tokens), and then within each class to object poses. Another limitation is the challenge in effectively distinguishing pose from orientation during clustering, especially in the case of articulated objects. Additionally, the absence of camera poses requires one to manually set the parameters to obtain explicit 3D from our generated views. This issue could in theory be addressed by estimating a 3D template from average DINOv2 features at the cluster centroids. We leave this for future exploration. Lastly, there is room for improvement in the consistency of the generated multi-view data.


{
    \small
    \bibliographystyle{ieeenat_fullname}
    \bibliography{main}
}

\appendix
\setcounter{section}{0} 

\section{Additional Implementation Details}
As stated in the main paper, during training, we use the hyperparameters outlined in DDIM~\cite{song2020denoising} and follow the code from the official implementation~\footnote{\url{https://github.com/ermongroup/ddim}}. 
We adopt the U-Net architecture from this source~\footnote{\url{https://github.com/filipbasara0/simple-diffusion/tree/main/model}}. 
To make the architecture pose-conditioned, we incorporate learnable pose embeddings into the time embedding. 
We train our diffusion models on the CompCars~\cite{compcars}, SD Armchairs, and SD Mugs datasets (the latter is described in the next section) for $10^{6}$, $5\times10^{5}$, and $7\times10^{5}$ iterations, respectively. 
All models are trained using four NVIDIA GeForce RTX 3090 GPUs. 
When applied, the cross-frame attention is used throughout all time steps. 
We will release the code upon the acceptance of our paper.

\section{Additional Experimental Results}

In this section, we report some additional results of \methodName that could not be included in the main paper due to the page number limit. 

\begin{figure*}
\begin{center}
    \centering
    \captionsetup{type=figure}
        \begin{tabular}{@{}c
        @{\hspace{0.5mm}}c@{\hspace{0.5mm}}c@{\hspace{0.5mm}}c@{\hspace{0.5mm}}c@{\hspace{0.5mm}}c@{\hspace{0.5mm}}c@{\hspace{0.5mm}}} 
            Reference & \multicolumn{6}{c}{Generated Novel Views}  \\
            {\includegraphics[width=0.124\linewidth]{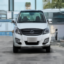}} &
            {\includegraphics[width=0.124\linewidth]{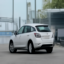}} & {\includegraphics[width=0.124\linewidth]{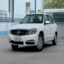}} & 
            {\includegraphics[width=0.124\linewidth]{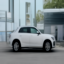}} & 
            {\includegraphics[width=0.124\linewidth]{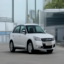}} &
            {\includegraphics[width=0.124\linewidth]{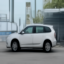}} & 
            {\includegraphics[width=0.124\linewidth]{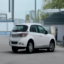}} \\
            {\includegraphics[width=0.124\linewidth]{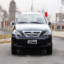}} &
            {\includegraphics[width=0.124\linewidth]{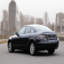}} & {\includegraphics[width=0.124\linewidth]{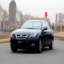}} & 
            {\includegraphics[width=0.124\linewidth]{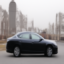}} & 
            {\includegraphics[width=0.124\linewidth]{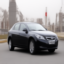}} &
            {\includegraphics[width=0.124\linewidth]{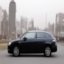}} & 
            {\includegraphics[width=0.124\linewidth]{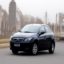}} \\
            {\includegraphics[width=0.124\linewidth]{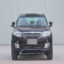}} &
            {\includegraphics[width=0.124\linewidth]{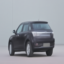}} & {\includegraphics[width=0.124\linewidth]{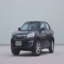}} & 
            {\includegraphics[width=0.124\linewidth]{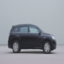}} & 
            {\includegraphics[width=0.124\linewidth]{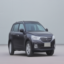}} &
            {\includegraphics[width=0.124\linewidth]{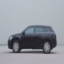}} & 
            {\includegraphics[width=0.124\linewidth]{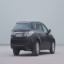}} \\
            {\includegraphics[width=0.124\linewidth]{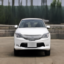}} &
            {\includegraphics[width=0.124\linewidth]{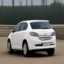}} & {\includegraphics[width=0.124\linewidth]{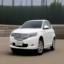}} & 
            {\includegraphics[width=0.124\linewidth]{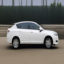}} & 
            {\includegraphics[width=0.124\linewidth]{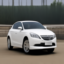}} &
            {\includegraphics[width=0.124\linewidth]{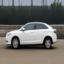}} & 
            {\includegraphics[width=0.124\linewidth]{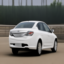}} \\
            {\includegraphics[width=0.124\linewidth]{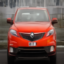}} &
            {\includegraphics[width=0.124\linewidth]{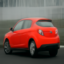}} & {\includegraphics[width=0.124\linewidth]{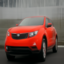}} & 
            {\includegraphics[width=0.124\linewidth]{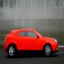}} & 
            {\includegraphics[width=0.124\linewidth]{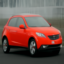}} &
            {\includegraphics[width=0.124\linewidth]{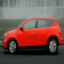}} & 
            {\includegraphics[width=0.124\linewidth]{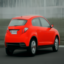}} \\
            & & & & & & \\
            {\includegraphics[width=0.124\linewidth]{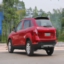}} &
            {\includegraphics[width=0.124\linewidth]{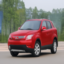}} & {\includegraphics[width=0.124\linewidth]{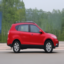}} & 
            {\includegraphics[width=0.124\linewidth]{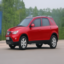}} & 
            {\includegraphics[width=0.124\linewidth]{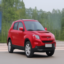}} &
            {\includegraphics[width=0.124\linewidth]{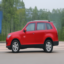}} & 
            {\includegraphics[width=0.124\linewidth]{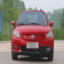}} \\
            {\includegraphics[width=0.124\linewidth]{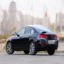}} &
            {\includegraphics[width=0.124\linewidth]{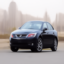}} & {\includegraphics[width=0.124\linewidth]{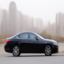}} & 
            {\includegraphics[width=0.124\linewidth]{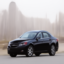}} & 
            {\includegraphics[width=0.124\linewidth]{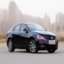}} &
            {\includegraphics[width=0.124\linewidth]{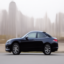}} & 
            {\includegraphics[width=0.124\linewidth]{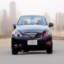}} \\
            {\includegraphics[width=0.124\linewidth]{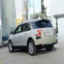}} &
            {\includegraphics[width=0.124\linewidth]{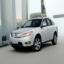}} & {\includegraphics[width=0.124\linewidth]{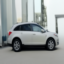}} & 
            {\includegraphics[width=0.124\linewidth]{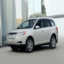}} & 
            {\includegraphics[width=0.124\linewidth]{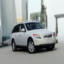}} &
            {\includegraphics[width=0.124\linewidth]{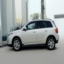}} & 
            {\includegraphics[width=0.124\linewidth]{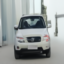}} \\
            {\includegraphics[width=0.124\linewidth]{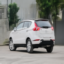}} &
            {\includegraphics[width=0.124\linewidth]{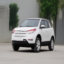}} & {\includegraphics[width=0.124\linewidth]{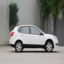}} & 
            {\includegraphics[width=0.124\linewidth]{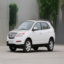}} & 
            {\includegraphics[width=0.124\linewidth]{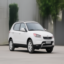}} &
            {\includegraphics[width=0.124\linewidth]{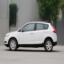}} & 
            {\includegraphics[width=0.124\linewidth]{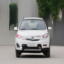}} \\
            {\includegraphics[width=0.124\linewidth]{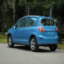}} &
            {\includegraphics[width=0.124\linewidth]{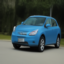}} & {\includegraphics[width=0.124\linewidth]{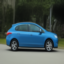}} & 
            {\includegraphics[width=0.124\linewidth]{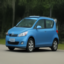}} & 
            {\includegraphics[width=0.124\linewidth]{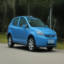}} &
            {\includegraphics[width=0.124\linewidth]{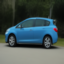}} & 
            {\includegraphics[width=0.124\linewidth]{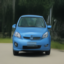}} \\

        \end{tabular}
    \captionof{figure}{
    Generated cars with \methodName for different reference poses (front view and half-back view).\label{fig:our_cars_1}}
\end{center}
\end{figure*}

\begin{figure*}
\begin{center}
    \centering
    \captionsetup{type=figure}
        \begin{tabular}{@{}c
        @{\hspace{0.5mm}}c@{\hspace{0.5mm}}c@{\hspace{0.5mm}}c@{\hspace{0.5mm}}c@{\hspace{0.5mm}}c@{\hspace{0.5mm}}c@{\hspace{0.5mm}}} 
            Reference   & \multicolumn{6}{c}{Generated Novel Views}  \\

            {\includegraphics[width=0.124\linewidth]{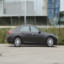}} &
            {\includegraphics[width=0.124\linewidth]{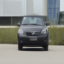}} & {\includegraphics[width=0.124\linewidth]{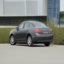}} & 
            {\includegraphics[width=0.124\linewidth]{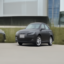}} & 
            {\includegraphics[width=0.124\linewidth]{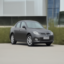}} &
            {\includegraphics[width=0.124\linewidth]{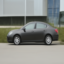}} & 
            {\includegraphics[width=0.124\linewidth]{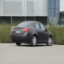}} \\
            {\includegraphics[width=0.124\linewidth]{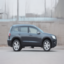}} &
            {\includegraphics[width=0.124\linewidth]{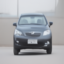}} & {\includegraphics[width=0.124\linewidth]{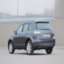}} & 
            {\includegraphics[width=0.124\linewidth]{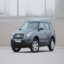}} & 
            {\includegraphics[width=0.124\linewidth]{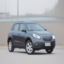}} &
            {\includegraphics[width=0.124\linewidth]{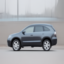}} & 
            {\includegraphics[width=0.124\linewidth]{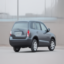}} \\
            {\includegraphics[width=0.124\linewidth]{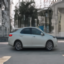}} &
            {\includegraphics[width=0.124\linewidth]{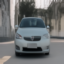}} & {\includegraphics[width=0.124\linewidth]{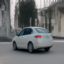}} & 
            {\includegraphics[width=0.124\linewidth]{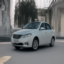}} & 
            {\includegraphics[width=0.124\linewidth]{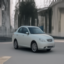}} &
            {\includegraphics[width=0.124\linewidth]{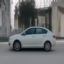}} & 
            {\includegraphics[width=0.124\linewidth]{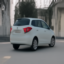}} \\
            {\includegraphics[width=0.124\linewidth]{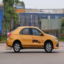}} &
            {\includegraphics[width=0.124\linewidth]{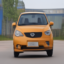}} & {\includegraphics[width=0.124\linewidth]{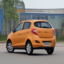}} & 
            {\includegraphics[width=0.124\linewidth]{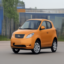}} & 
            {\includegraphics[width=0.124\linewidth]{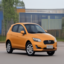}} &
            {\includegraphics[width=0.124\linewidth]{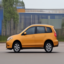}} & 
            {\includegraphics[width=0.124\linewidth]{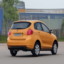}} \\

        \end{tabular}
    \captionof{figure}{
    Generated cars with \methodName for the case of the side-view as the reference pose.\label{fig:our_cars_2}}
\end{center}
\end{figure*}

First, we show further samples of generated novel views on the CompCars dataset (see Figures~\ref{fig:our_cars_1} and \ref{fig:our_cars_2}). Notice that the model is able to generate plausible novel views from different reference poses. 
We observe that the greater the visibility of features in the reference image, the lower the variability in shape/texture observed in the generated views, resulting in improved consistency that aligns with our expectations.

\begin{figure}
\begin{center}
    \centering
    \captionsetup{type=figure}
        \begin{tabular}{@{}r@{\hspace{1.5mm}}c@{\hspace{0.5mm}}c@{\hspace{0.5mm}}c@{\hspace{0.5mm}}c@{\hspace{0.5mm}}c@{\hspace{0.5mm}}c@{\hspace{0.5mm}}c@{\hspace{0.5mm}}c@{}} 
            Reference & \multicolumn{4}{c}{Generated Novel Views}  \\
            {\includegraphics[width=0.2\linewidth]{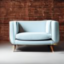}} &
            {\includegraphics[width=0.2\linewidth]{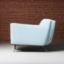}} & {\includegraphics[width=0.2\linewidth]{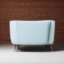}} & 
            {\includegraphics[width=0.2\linewidth]{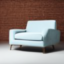}} & 
            {\includegraphics[width=0.2\linewidth]{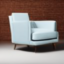}} &  \\
            {\includegraphics[width=0.2\linewidth]{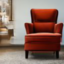}} &
            {\includegraphics[width=0.2\linewidth]{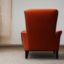}} & {\includegraphics[width=0.2\linewidth]{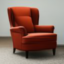}} & 
            {\includegraphics[width=0.2\linewidth]{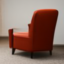}} & 
            {\includegraphics[width=0.2\linewidth]{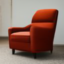}} &  \\
            {\includegraphics[width=0.2\linewidth]{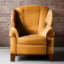}} &
            {\includegraphics[width=0.2\linewidth]{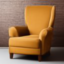}} & {\includegraphics[width=0.2\linewidth]{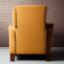}} & 
            {\includegraphics[width=0.2\linewidth]{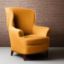}} & 
            {\includegraphics[width=0.2\linewidth]{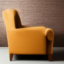}} &  \\
            {\includegraphics[width=0.2\linewidth]{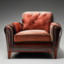}} &
            {\includegraphics[width=0.2\linewidth]{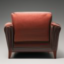}} & {\includegraphics[width=0.2\linewidth]{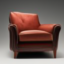}} & 
            {\includegraphics[width=0.2\linewidth]{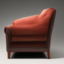}} & 
            {\includegraphics[width=0.2\linewidth]{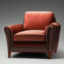}} &  \\
            & & & & \\
            {\includegraphics[width=0.2\linewidth]{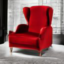}} &
            {\includegraphics[width=0.2\linewidth]{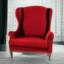}} & {\includegraphics[width=0.2\linewidth]{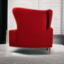}} & 
            {\includegraphics[width=0.2\linewidth]{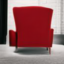}} & 
            {\includegraphics[width=0.2\linewidth]{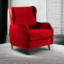}} &  \\
            {\includegraphics[width=0.2\linewidth]{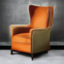}} &
            {\includegraphics[width=0.2\linewidth]{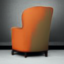}} & {\includegraphics[width=0.2\linewidth]{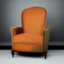}} & 
            {\includegraphics[width=0.2\linewidth]{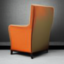}} & 
            {\includegraphics[width=0.2\linewidth]{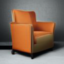}} &  \\
            {\includegraphics[width=0.2\linewidth]{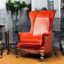}} &
            {\includegraphics[width=0.2\linewidth]{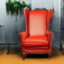}} & {\includegraphics[width=0.2\linewidth]{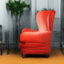}} & 
            {\includegraphics[width=0.2\linewidth]{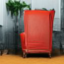}} & 
            {\includegraphics[width=0.2\linewidth]{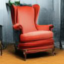}} &  \\
            {\includegraphics[width=0.2\linewidth]{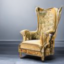}} &
            {\includegraphics[width=0.2\linewidth]{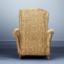}} & {\includegraphics[width=0.2\linewidth]{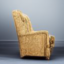}} & 
            {\includegraphics[width=0.2\linewidth]{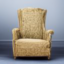}} & 
            {\includegraphics[width=0.2\linewidth]{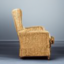}} &  \\
            & & & & \\
            {\includegraphics[width=0.2\linewidth]{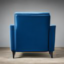}} &
            {\includegraphics[width=0.2\linewidth]{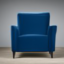}} & {\includegraphics[width=0.2\linewidth]{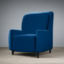}} & 
            {\includegraphics[width=0.2\linewidth]{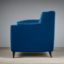}} & 
            {\includegraphics[width=0.2\linewidth]{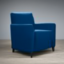}} &  \\
            {\includegraphics[width=0.2\linewidth]{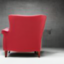}} &
            {\includegraphics[width=0.2\linewidth]{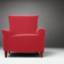}} & {\includegraphics[width=0.2\linewidth]{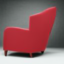}} & 
            {\includegraphics[width=0.2\linewidth]{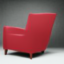}} & 
            {\includegraphics[width=0.2\linewidth]{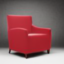}} &  \\
            {\includegraphics[width=0.2\linewidth]{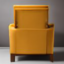}} &
            {\includegraphics[width=0.2\linewidth]{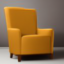}} & {\includegraphics[width=0.2\linewidth]{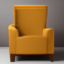}} & 
            {\includegraphics[width=0.2\linewidth]{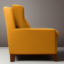}} & 
            {\includegraphics[width=0.2\linewidth]{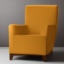}} &  \\
            {\includegraphics[width=0.2\linewidth]{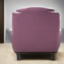}} &
            {\includegraphics[width=0.2\linewidth]{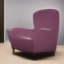}} & {\includegraphics[width=0.2\linewidth]{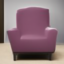}} & 
            {\includegraphics[width=0.2\linewidth]{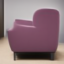}} & 
            {\includegraphics[width=0.2\linewidth]{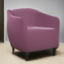}} &  \\
        \end{tabular}
    \captionof{figure}{
    Additional examples of novel views generated by \methodName on the SD Armchairs dataset from different reference poses (front, half-front, and back views).\label{fig:armchairs_generated_diff_ref_poses}}
\end{center}
\end{figure}

We also show additional samples from our model trained on the SD Armchairs dataset in Figure~\ref{fig:armchairs_generated_diff_ref_poses}. Again, regardless of the reference pose, the model is able to generate plausible novel views.

\begin{figure}
\begin{center}
    \centering
    \captionsetup{type=figure}
        \begin{tabular}{@{}r@{\hspace{1.5mm}}c@{\hspace{0.5mm}}c@{\hspace{0.5mm}}c@{\hspace{0.5mm}}c@{\hspace{0.5mm}}c@{\hspace{0.5mm}}c@{\hspace{0.5mm}}c@{}} 
            Reference & \multicolumn{3}{c}{Generated Novel Views}  \\
            {\includegraphics[width=0.2\linewidth]{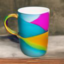}} &
            {\includegraphics[width=0.2\linewidth]{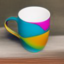}} & 
            {\includegraphics[width=0.2\linewidth]{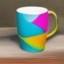}} & 
            {\includegraphics[width=0.2\linewidth]{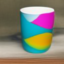}}   \\
            {\includegraphics[width=0.2\linewidth]{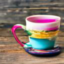}} &
            {\includegraphics[width=0.2\linewidth]{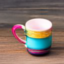}} & 
            {\includegraphics[width=0.2\linewidth]{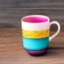}} & 
            {\includegraphics[width=0.2\linewidth]{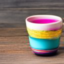}}   \\
            {\includegraphics[width=0.2\linewidth]{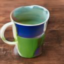}} &
            {\includegraphics[width=0.2\linewidth]{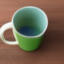}} & 
            {\includegraphics[width=0.2\linewidth]{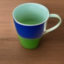}} & 
            {\includegraphics[width=0.2\linewidth]{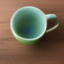}}   \\
            {\includegraphics[width=0.2\linewidth]{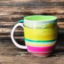}} &
            {\includegraphics[width=0.2\linewidth]{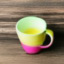}} & 
            {\includegraphics[width=0.2\linewidth]{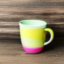}} & 
            {\includegraphics[width=0.2\linewidth]{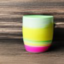}}   \\
            & & &  \\
            {\includegraphics[width=0.2\linewidth]{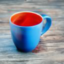}} &
            {\includegraphics[width=0.2\linewidth]{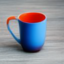}} & 
            {\includegraphics[width=0.2\linewidth]{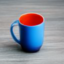}} & 
            {\includegraphics[width=0.2\linewidth]{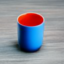}}   \\
            {\includegraphics[width=0.2\linewidth]{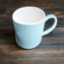}} &
            {\includegraphics[width=0.2\linewidth]{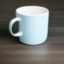}} & 
            {\includegraphics[width=0.2\linewidth]{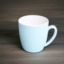}} & 
            {\includegraphics[width=0.2\linewidth]{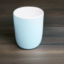}}   \\
            {\includegraphics[width=0.2\linewidth]{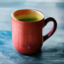}} &
            {\includegraphics[width=0.2\linewidth]{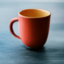}} & 
            {\includegraphics[width=0.2\linewidth]{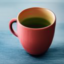}} & 
            {\includegraphics[width=0.2\linewidth]{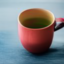}}   \\ 
            {\includegraphics[width=0.2\linewidth]{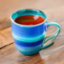}} &
            {\includegraphics[width=0.2\linewidth]{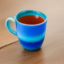}} & 
            {\includegraphics[width=0.2\linewidth]{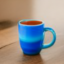}} & 
            {\includegraphics[width=0.2\linewidth]{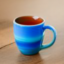}}   \\ 
            & & & \\
            {\includegraphics[width=0.2\linewidth]{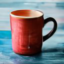}} &
            {\includegraphics[width=0.2\linewidth]{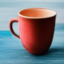}} & 
            {\includegraphics[width=0.2\linewidth]{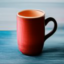}} & 
            {\includegraphics[width=0.2\linewidth]{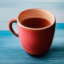}}  \\
            {\includegraphics[width=0.2\linewidth]{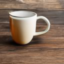}} &
            {\includegraphics[width=0.2\linewidth]{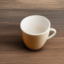}} & 
            {\includegraphics[width=0.2\linewidth]{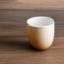}} & 
            {\includegraphics[width=0.2\linewidth]{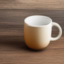}}   \\
            {\includegraphics[width=0.2\linewidth]{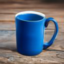}} &
            {\includegraphics[width=0.2\linewidth]{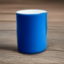}} & 
            {\includegraphics[width=0.2\linewidth]{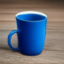}} & 
            {\includegraphics[width=0.2\linewidth]{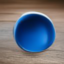}}  \\
            {\includegraphics[width=0.2\linewidth]{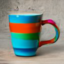}} &
            {\includegraphics[width=0.2\linewidth]{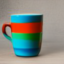}} & 
            {\includegraphics[width=0.2\linewidth]{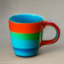}} & 
            {\includegraphics[width=0.2\linewidth]{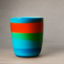}}  \\
        \end{tabular}
    \captionof{figure}{
    Additional examples of novel views generated by \methodName on the SD Mugs dataset from different reference poses (side views and top-side view). Note that columns do not necessarily represent the same pose cluster.\label{fig:mugs_generated_diff_ref_poses}}
\end{center}
\end{figure}

To emphasize the robustness of our pipeline to different categories, we generate another synthetic dataset from Stable Diffusion. 
We follow the same steps as for the SD Armchairs dataset, but this time we used the category of ``a single colorful mug''. We call the resulting dataset \textit{SD Mugs}. 
The qualitative results of novel view synthesis on the SD Mugs dataset is shown in Figure~\ref{fig:mugs_generated_diff_ref_poses}, while some quantitative data is reported in Table~\ref{table:FID}. Notice, that due to the handle, mugs have a fundamentally different topology compared to cars and armchairs. For instance, \cite{monnier2022unicorn} are restricted to modelling objects that are homeomorphic to spheres. \methodName does not have such a restriction and is able to work for diverse object topologies.

\begin{figure*}
\begin{center}
    \centering
    \captionsetup{type=figure}
        \begin{tabular}{@{}c@{\hspace{0.5mm}}c@{\hspace{0.5mm}}c@{\hspace{0.5mm}}c@{\hspace{0.5mm}}c@{\hspace{0.5mm}}c@{\hspace{0.5mm}}c@{\hspace{0.5mm}}c@{\hspace{0.5mm}}} 

            {\includegraphics[width=0.124\linewidth]{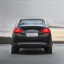}} &
            {\includegraphics[width=0.124\linewidth]{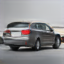}} & {\includegraphics[width=0.124\linewidth]{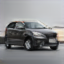}} & 
            {\includegraphics[width=0.124\linewidth]{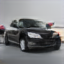}} & 
            {\includegraphics[width=0.124\linewidth]{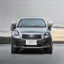}} &
            {\includegraphics[width=0.124\linewidth]{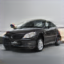}} & 
            {\includegraphics[width=0.124\linewidth]{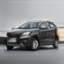}}& 
            {\includegraphics[width=0.124\linewidth]{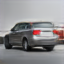}} \\
            {\includegraphics[width=0.124\linewidth]{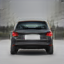}} &
            {\includegraphics[width=0.124\linewidth]{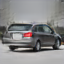}} & {\includegraphics[width=0.124\linewidth]{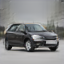}} & 
            {\includegraphics[width=0.124\linewidth]{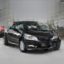}} & 
            {\includegraphics[width=0.124\linewidth]{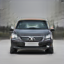}} &
            {\includegraphics[width=0.124\linewidth]{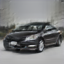}} & 
            {\includegraphics[width=0.124\linewidth]{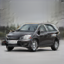}}& 
            {\includegraphics[width=0.124\linewidth]{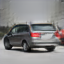}} \\
            {\includegraphics[width=0.124\linewidth]{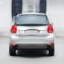}} &
            {\includegraphics[width=0.124\linewidth]{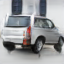}} & {\includegraphics[width=0.124\linewidth]{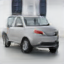}} & 
            {\includegraphics[width=0.124\linewidth]{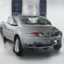}} & 
            {\includegraphics[width=0.124\linewidth]{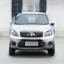}} &
            {\includegraphics[width=0.124\linewidth]{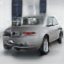}} & 
            {\includegraphics[width=0.124\linewidth]{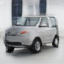}}& 
            {\includegraphics[width=0.124\linewidth]{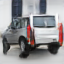}} \\
            {\includegraphics[width=0.124\linewidth]{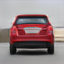}} &
            {\includegraphics[width=0.124\linewidth]{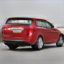}} & {\includegraphics[width=0.124\linewidth]{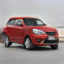}} & 
            {\includegraphics[width=0.124\linewidth]{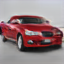}} & 
            {\includegraphics[width=0.124\linewidth]{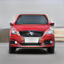}} &
            {\includegraphics[width=0.124\linewidth]{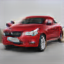}} & 
            {\includegraphics[width=0.124\linewidth]{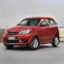}}& 
            {\includegraphics[width=0.124\linewidth]{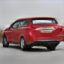}} \\
            {\includegraphics[width=0.124\linewidth]{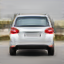}} &
            {\includegraphics[width=0.124\linewidth]{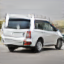}} & {\includegraphics[width=0.124\linewidth]{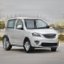}} & 
            {\includegraphics[width=0.124\linewidth]{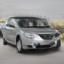}} & 
            {\includegraphics[width=0.124\linewidth]{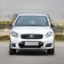}} &
            {\includegraphics[width=0.124\linewidth]{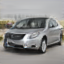}} & 
            {\includegraphics[width=0.124\linewidth]{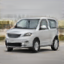}}& 
            {\includegraphics[width=0.124\linewidth]{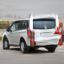}} \\

        \end{tabular}
    \captionof{figure}{For comparison, 
    $360^{\circ}$ generated cars using GIRAFFE~\cite{GIRAFFE}.\label{fig:giraffe_compcars}}
\end{center}
\end{figure*}

Lastly, we also show some samples from  GIRAFFE~\ref{fig:giraffe_compcars}, which is also a generative model to create novel views of objects. We took the pretrained GIRAFFE model on the CompCars dataset and generated $360^{\circ}$ views. The generated views do not exhibit a strong multi-view consistency, as both the cars' identities (shapes) and their textures change with the viewpoint. This is a well know problem for 3D-aware Generative Adversarial Networks (GANs), ``as none of them can preserve strict multi-view consistency, partially on account of the usage of a 2D upsampler and lack of explicit 3D supervision'' (as quoted from \cite{lift3D2023CVPR}).

\begin{table}[t]
\centering
\begin{tabular}{c l r }
\toprule
Dataset & Method & FID $\downarrow$ \\ \hline
\multirow{3}{*}{SD Mugs} & Ours - no MV & 5.228  \\
& Ours - MV, w/o HAG & 11.666 \\
& Ours - MV, w/ HAG &  15.487 \\ \hline
\end{tabular}
\caption{The FID scores for SD Mugs for various versions of \methodName. MV refers to multi-view.} \label{table:FID}
\end{table}

\section{Failure Cases}
\begin{figure*}
\begin{center}
    \centering
    \captionsetup{type=figure}
        \begin{tabular}{@{}c@{\hspace{0.5mm}}c@{\hspace{0.5mm}}c@{\hspace{0.5mm}}c@{\hspace{0.5mm}}c@{\hspace{0.5mm}}c@{\hspace{0.5mm}}c@{\hspace{0.5mm}}c@{\hspace{0.5mm}}} 
            Reference   & \multicolumn{7}{c}{Generated Novel Views}  \\
            {\includegraphics[width=0.124\linewidth]{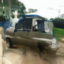}} &
            {\includegraphics[width=0.124\linewidth]{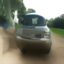}} & {\includegraphics[width=0.124\linewidth]{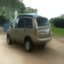}} & 
            {\includegraphics[width=0.124\linewidth]{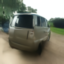}} & 
            {\includegraphics[width=0.124\linewidth]{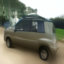}} &
            {\includegraphics[width=0.124\linewidth]{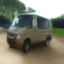}} & 
            {\includegraphics[width=0.124\linewidth]{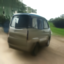}}& 
            {\includegraphics[width=0.124\linewidth]{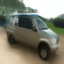}} \\
            {\includegraphics[width=0.124\linewidth]{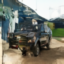}} &
            {\includegraphics[width=0.124\linewidth]{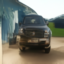}} & {\includegraphics[width=0.124\linewidth]{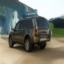}} & 
            {\includegraphics[width=0.124\linewidth]{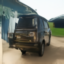}} & 
            {\includegraphics[width=0.124\linewidth]{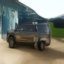}} &
            {\includegraphics[width=0.124\linewidth]{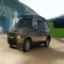}} & 
            {\includegraphics[width=0.124\linewidth]{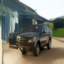}}& 
            {\includegraphics[width=0.124\linewidth]{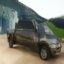}} \\

        \end{tabular}
    \captionof{figure}{
    Failure cases for the generated cars.\label{fig:failure_case}}
\end{center}
\end{figure*}

We observe that the quality of the generated views highly depends on the quality of the reference pose. If the generated reference pose is not a valid image, then the generated views lack both realism and multi-view consistency. This can be clearly observed in Figure~\ref{fig:failure_case}, where the generated novel views for two invalid reference images are both unrealistic and lack multi-view consistency.

\end{document}